# An Empirical Analysis of the Efficacy of Different Sampling Techniques for Imbalanced Classification

Asif Newaz [1]*, Shahriar Hassan [2], Farhan Shahriyar Haq [3]

[1,2] Department of Electrical and Electronic Engineering,

Islamic University of Technology, Gazipur, Bangladesh;

[3] Robi Axiata Limited, Bangladesh;

**Email:** eee.asifnewaz@iut-dhaka.edu [1], shahriarhassan@iut-dhaka.edu [2],

farhanshahriyar@iut-dhaka.edu [3]

**\* Corresponding Author:**

**Address:** Department of Electrical and Electronic Engineering, Islamic University of Technology, Gazipur-1704, Bangladesh.

**Email:** eee.asifnewaz@iut-dhaka.edu

**Contact:** +8801880841119

# Abstract

Learning from imbalanced data is a challenging task. Standard classification algorithms tend to perform poorly when trained on imbalanced data. Some special strategies need to be adopted, either by modifying the data distribution or by redesigning the underlying classification algorithm to achieve desirable performance. The prevalence of imbalance in real-world datasets has led to the creation of a multitude of strategies for the class imbalance issue. However, not all the strategies are useful or provide good performance in different imbalance scenarios. There are numerous approaches to dealing with imbalanced data, but the efficacy of such techniques or an experimental comparison among those techniques has not been conducted. In this study, we present a comprehensive analysis of 26 popular sampling techniques to understand their effectiveness in dealing with imbalanced data. Rigorous experiments have been conducted on 50 datasets with different degrees of imbalance to thoroughly investigate the performance of these techniques. A detailed discussion of the advantages and limitations of the techniques, as well as how to overcome such limitations, has been presented. We identify some critical factors that affect the sampling strategies and provide recommendations on how to choose an appropriate sampling technique for a particular application.

# 1. Introduction

Class imbalance refers to having an unequal number of samples in different classes in the dataset. The objective of the classification algorithms is to predict each class correctly. However, if one class is underrepresented in the data, the prediction becomes biased towards the majority class. Real-world datasets are often imbalanced and generally, it is the positive class that has a relatively small number of instances. However, identifying these positive instances is usually more important. So, the presence of imbalance in the data poses a major challenge in the classification tasks. The imbalance classification problem has drawn a lot of attention from researchers over the years [1]. Many different techniques have been proposed to address this problem. These techniques have been proven to be quite useful in imbalance classification tasks [2–5].

The problem with imbalanced data is that the standard classification algorithms are not designed to consider class imbalance. They are built on the assumption of an equal number of instances in different classes. The classifiers are trained to minimize the number of wrong predictions irrespective of the class. Due to this accuracy-oriented design, the classifier fails to differentiate between the minority and majority class instances. This leads to a bias towards the majority class. So, when there is an uneven class distribution, the classifier naturally fails to provide satisfactory performance. To achieve reasonable performance, some modifications to the algorithms or the data are essential. The techniques developed to deal with class imbalance either modify the data on which the classifier is to be trained or modify the algorithms to adapt themselves so that they can learn from imbalanced data. These techniques can be divided into four categories.

1. Data-level modifications: This is the most popular approach in imbalanced classification tasks. The idea is to balance the dataset by introducing new minority class samples or by removing samples from the majority class. The classifier is then trained on the balanced dataset. This approach is generally known as "sampling". The success of the sampling techniques depends largely on how the data is modified to obtain balance. It can be done heuristically or non-heuristically. Many algorithms have been proposed on how to generate new samples or eliminate samples from the dataset. These can be broadly classified as undersampling, oversampling, and hybrid approaches.
2. Algorithm-level modifications: The existing classification algorithms can be adapted to bias the learning towards the minority class. In this way, the classifiers are enabled to learn from imbalanced data without any kind of data-level modification.
3. Cost-sensitive learning: The standard classification algorithms assign equal misclassification costs to both majority and minority class instances. The classifier is then trained to minimize the overall cost. In the cost-sensitive learning framework, this is modified by assigning a higher cost to the misclassification of minority class instances. This way, the classifier is modified to impart more importance to the minority class samples while learning.

4. Ensemble-based approach: Although ensemble algorithms are usually considered superior, they remain susceptible to skewed data distribution. In this approach, the sampling techniques or cost-sensitive learning are combined with the ensemble learning framework to address the class imbalance scenario. Here, the sampling techniques are incorporated into the bootstrapping process, or cost-sensitive learning is incorporated into the training process to modify the ensemble algorithms to learn from imbalanced data.

Table 1 summarizes the popular approaches in the imbalanced domain.

Table 1: Popular techniques in Imbalanced learning

| Type | Techniques |
| --- | --- |
| Data-level modification | 1. Oversampling:<br>   - Random Oversampling (ROS)<br>   - Synthetic Minority Oversampling (SMOTE) [6]<br>   - Adaptive SMOTE (ADASYN) [7]<br>   - Borderline-SMOTE [8]<br>   - Safe-level-SMOTE [9]<br>   - CURE-SMOTE [10]<br>   - Density-Based SMOTE (DBSMOTE) [11]<br>   - Majority Weighted Minority Oversampling Technique (MWMOTE) [12]<br>   - Random Walk Oversampling (RWO-Sampling) [13]<br>   - Borderline Kernel-based Oversampling (BKS) [14]<br>   - NEATER [15]<br>   - wRACOG [16]<br>2. Undersampling:<br>   - Random Undersampling (RUS)<br>   - Cluster-Based Evolutionary Undersampling (CBEUS) [17]<br>   - Ant-Colony based Undersampling (ACOSampling) [18]<br>   - Instance Hardness Threshold (IHT) [19]<br>   - Inverse Random Undersampling (IRUS) [20]<br>   - Diversified Sensitivity Undersampling (DSUS) [21]<br>   - Tomek-links (TL) [22] |
|  | - Condensed Nearest Neighbor rule (CNN) [23] |
|  | - Edited Nearest Neighbor rule (ENN) [24] |
|  | - Repeated Edited Nearest Neighbor rule (R-ENN) [25] |
|  | - Neighborhood Cleaning Rule (NC) [26] |
|  | - Near-Miss approaches [27] |
|  | 3. Hybrid Sampling: |
|  | - One-Sided Selection (OSS) [28] |
|  | - SMOTE-Tomek [29] |
|  | - SMOTE-ENN |
|  | - Random Balance [30] |
|  | - SMOTE-RUS-NC [31] |
| Algorithm-level modifications | - SVM and their variations [32–34] |
|  | - Fuzzy SVM (FSVM) [35] |
|  | - Decision Trees [36] |
|  | - Bayesian classifiers [37] |
|  | - ANNs [38] |
|  | - Kernel Machines [39] |
| Cost-sensitive learning | - MetaCost [40] |
|  | - Cost-sensitive Decision Trees [41] |
|  | - Cost-sensitive SVM [42] |
| Ensemble-based approach | 1. Bagging Ensembles: |
|  | - OverBagging |
|  | - UnderBagging |
|  | - SMOTEBagging [43] |
|  | - Roughly Balanced Bagging [44] |
|  | - UnderOverBagging [43] |
|  | - Balanced Random Forest [45] |
|  | - Random Balance Bagging [30] |
|  | 2. Boosting Ensembles: |
|  | - SMOTEBoost [46] |
|  | - RUSBoost [47] |
|  | - BalancedBoost [48] |
|  | - Evolutionary Undersampling Boosting (EUSBoost) [49] |

3. Hybrid Ensembles:
   - EasyEnsemble [50]
   - HardEnsemble [51]
4. Cost-sensitive:
   - AdaCost [52]
   - RareBoost [53]

While there has been a development of a plethora of approaches to deal with imbalance data, there has been only a limited amount of research analyzing the effectiveness of these techniques [54–56]. There are a vast number of proposals, but an experimental comparison among those techniques has not been conducted. Not every technique works well in all scenarios. While one algorithm might perform quite well in one task but might fail to do well in another. So, the question arises: which algorithm is more effective in which scenario? Also, why and when does a particular algorithm succeed or fail in a classification task? Since there exist a sheer number of options to choose from, which one would be more suitable for a particular task? In this study, rigorous experiments have been conducted to thoroughly investigate the performance of these techniques. A comparative analysis among the approaches is provided, while also exploring the advantages and limitations of the techniques.

The contribution of this research article is summarized below.
- First, we look into the overall performance of the techniques on a wide variety of datasets. This provides a basic idea of their performance and a comparison among the techniques.
- Next, we look into how these techniques tackle the imbalance scenario and how effective the process is on a wide range of datasets.
- We analyze how different degrees of imbalance and other data characteristics affect the overall performance.
- We particularly examine severely imbalanced scenarios as they are more difficult to learn. We study the performance of different techniques to identify which algorithms produce acceptable performance and hypothesize how better performance can be achieved in those cases.
- We investigate the stability of these techniques and identify which are high-variance strategies. We also analyze the reasons behind them, and how that can be reduced.
- We investigate the time complexity of these techniques.
- We discuss the advantages and drawbacks of the strategies based on their performance and try to come to a general conclusion on which algorithms are more suitable for which scenarios.

The remaining article is structured as follows: In Section 2, related research works are discussed. In Section 3, some critical issues in the imbalanced learning domain that need to be taken care of are highlighted. In Section 4, the experimental design is thoroughly discussed along with the datasets utilized for the experimentation. The algorithms utilized in this study are described in Section 5. In Section 6, we present the results. In Section 7, a detailed discussion and comparative analysis have been provided. Finally, the article is concluded in Section 8 with a discussion on the limitations of this study as well as some of the open research issues on imbalanced learning.

## 2. Literature Review

There have been a substantial number of proposals over the years on how to deal with imbalance scenarios. However, the number of research articles discussing the efficacy of the techniques or providing a comparison among these techniques is quite limited. There have also been several review articles on the techniques used in the imbalanced domain and their applications [57-59]. These articles mainly focus on the use of different techniques, not an investigation into the effectiveness of those strategies. The most relevant papers that investigate the applicability of these techniques are summarized below.

Blagus et al. [54] provided an experimental analysis of the applicability of the SMOTE algorithm on high-dimensional class imbalanced data. The authors conclude in their research that SMOTE is efficient in low-dimensional settings but fails in high-dimensional data. A variable selection must be performed beforehand for SMOTE to have any effect on high-dimensional datasets. The authors also conduct experiments with different algorithms and conclude that SMOTE is not favorable for Discriminant Analysis (DA), even in low-dimensional settings. Other classifiers generally benefit from SMOTE in low-dimensional data.

Prati et al. [60] assessed the performance of five different approaches – ROS, SMOTE, Borderline-SMOTE, ADASYN, and MetaCost – on 22 real-world datasets. They experimented with different classifiers and reported that all the classifiers are affected by the imbalance. However, the Support Vector Machine (SVM) classifier is the least sensitive to imbalance. Based on the experiments, it was found that all five of the techniques mentioned above failed in the case of highly imbalanced scenarios. They also reported that the two variations of SMOTE, ADASYN and Borderline-SMOTE, did not display any particular advantage over SMOTE.

López et al. [56] investigated the data intrinsic characteristics to gain insight into the factors that hinder the performance of the classifiers when the data is skewed. Class overlapping, noisy data, borderline instances, small disjuncts, and data shift are identified as the factors for poor performance in the research article. The authors also conducted experiments to compare several strategies in the imbalanced domain. They tested several sampling techniques along with cost-sensitive classifiers. They also compared three classifiers – SVM, KNN, and Decision Trees (DT). They concluded in their study that SMOTE and SMOTE-ENN are the best sampling techniques among the ones tested. However, cost-sensitive SVM classifier usually outperforms those approaches. They also evaluated the performance of five ensemble approaches and found SMOTE-Bagging as the best performing strategy.

Branco et al. [59] reviewed some critical issues in the imbalanced domain. One of the most critical factors while working with imbalanced datasets is the performance metric. Appropriate metrics need to be chosen to rigorously evaluate the performance of the techniques. Metrics like accuracy provide over-optimistic results due to bias, while metrics like recall or precision only show part of the result. They fail to represent the overall performance of the classifiers. The authors discussed different metrics that can be utilized to measure performance correctly. They also reported other scenarios like regression or multi-label classification where data imbalance is a major issue.

Data dimensionality is another issue that further complicates the learning process from imbalanced data. Reducing the number of features generally improves the overall performance of the classifiers. Khoshgoftaar et al. [61] investigated the effect of feature selection on imbalanced data. The authors reported in their study that feature selection techniques perform significantly better when the data is balanced. Therefore, they suggested balancing the dataset prior to the application of feature selection methodologies to obtain desirable results. Yin et al. [62] introduced two novel approaches: Decomposition-based and Hellinger-distance-based feature selection strategies to deal with high-dimensional imbalanced data. A summary of different dimensionality reduction techniques for imbalanced learning can be found in[1].

Stefanowski et al. [55] studied data characteristics that affect the performance of the preprocessing algorithms. The authors identified several difficulty factors that deteriorate performance. They categorized the minority class instances as safe, borderline, rare, and outliers and claimed that distinguishing the instances this way can improve the performance of different techniques.

Weiss et al. [63] performed a comparative analysis between cost-sensitive learning and sampling techniques (ROS and RUS). They did not observe any clear winner between the two approaches. However, they reported that in the case of large datasets with more than 10,000 examples, the cost-sensitive learning algorithm outperformed the sampling techniques. They also conducted a comparison between oversampling and undersampling techniques and reported that there was no definite winner. They infer in their study that the effectiveness of the sampling approaches is highly dependent on the dataset.

However, they did not specify any specific characteristics of the data that affected the performance. Moreover, they compared the cost-sensitive algorithm with non-heuristic sampling techniques, but not with the more popular heuristic approaches.

Chen et al. [45] proposed two novel ensemble approaches: Balanced Random Forest (BRF) and Weighted Random Forest (WRF) in their article. Both algorithms are based on the Random Forest classifier. However, BRF incorporates RUS in the bootstrapping process, while WRF utilizes a cost-sensitive learning approach. The authors concluded in their study that there is no clear winner between these two approaches. However, BRF is computationally more efficient than WRF.

The findings of different research articles can be summarized as follows: All the classifiers are more or less affected by the imbalance, and the algorithms, by design, cannot handle the class imbalance problem. Data sampling is the most popular approach for imbalanced learning. However, there is no clear conclusion on whether oversampling or undersampling is more effective in dealing with imbalanced data. There are also contradictory reports on whether cost-sensitive learning is better than sampling techniques. Although some articles provide experimental results comparing several sampling techniques, the studies are quite limited as they are conducted on a limited number of datasets and strategies. An extensive experimental comparison among the techniques has not yet been conducted. Some studies have focused on identifying the data intrinsic characteristics that influence the learning process. However, an in-depth analysis of the properties, advantages, and drawbacks of the techniques has not been performed. This study is aimed at filling these gaps in research and providing a thorough look at the different ways to deal with imbalanced learning.

## 3. Learning from Imbalanced Data

In this section, we discuss some of the critical issues that need to be taken into consideration while working with imbalanced data.

### 3.1 Data Leakage

Data leakage is a major problem when designing predictive models. Machine learning models are first trained on a portion of the dataset (training data) and then tested on unseen data (validation or testing data) to properly evaluate how the model is performing. This splitting strategy is important to avoid bias and ensure generalization over a large population. However, if some information from the training dataset was available during the testing time, the model would generate over-optimistic results. To have an accurate understanding of how the prediction model would perform in the real world, the model needs to be tested on unseen data. Otherwise, the model is likely to fail in the application stage. This issue is often termed "data leakage", and it is a major pitfall when building predictive models. This is a serious issue but is also quite common when working with imbalanced data [64]. It occurs when the sampling is performed on the entire dataset before splitting it into training and test sets. This leads to two issues. Firstly, the original distribution of the data is lost. The dataset is no longer imbalanced. Testing on balanced data would not reflect the real scenario. Secondly, most of the sampling processes use some information from the original samples to generate new samples or select some samples from the dataset. So, if the dataset is split after resampling, some information will be leaked to the testing set, leading to data leakage.

In order to avoid such issues, some preventive measures need to be taken. The original data should be split into training and validation/test sets before resampling. Sampling algorithms should be applied only to the training dataset, and their performance should be tested on the validation set. Cross-validation is a popular strategy when building machine learning models to ensure generalization. To avoid data leakage during cross-validation, sampling should be performed on each of the training folds separately.

## 3.2   Data Shift

Data shift is another critical problem in imbalanced domain [65]. It arises when the distributions of the training and testing sets are different and can cause a degradation in the performance of the predictive model. It can be classified as intrinsic or induced. The intrinsic data shift occurs naturally when working with real-world datasets and usually has a small effect on the outcome. However, induced data shift commonly occurs when dealing with imbalanced data. This happens when a k-fold cross-validation strategy is utilized to produce the results. The problem with cross-validation is that it randomly splits the original data into k-different folds. While only one-fold is used for testing, the remaining folds are used in training. However, since the samples are selected randomly from the data to divide them into different folds, the distribution in these folds is not necessarily the same as the original data. Highly imbalanced data might exacerbate the situation. As the number of minority class samples is quite limited in highly imbalanced data, during random selection, a particular fold might not even contain any minority class instances. Testing the model on such a fold cannot evaluate the discriminating capability of the predictive model. As a result, this kind of selection bias often leads to impractical results.

To avoid such a scenario, a stratified cross-validation strategy should be utilized in imbalanced learning. This strategy attempts to maintain a similar distribution as the original data in different partitions, reducing the effect of induced data shift. However, it should be noted that data shift cannot be eliminated entirely.

## 3.3 Evaluation Metrics

Proper evaluation of a predictive model is crucial in the design of machine learning classifiers. There are many different metrics available to assess performance. However, proper metrics should be chosen depending on the application area. This is particularly important in the imbalanced domain, as some of the most popular metrics are not useful in imbalanced classification tasks.

Classification accuracy is the most popular metric to evaluate the performance of a classifier. However, this metric easily gets biased towards the majority class and provides over-optimistic results in imbalanced classification tasks. Sensitivity and specificity are two metrics that provide class-specific performance. However, independently these two metrics cannot represent the overall performance. To obtain a more precise conclusion, compound metrics need to be utilized. Balanced accuracy provides the arithmetic mean of sensitivity and specificity and is utilized in imbalanced learning. A more comprehensive measure can be obtained using the g-mean metric, which is the geometric mean of sensitivity and specificity. Another popular metric that is widely used in imbalanced learning is the area under the receiver operating characteristics curve (ROC-AUC score). It utilizes the True Positive Rate (TPR) and False Positive Rate (FPR) to obtain the ROC curve. These metrics are more suitable for imbalanced domain and should be utilized to properly assess the effectiveness of different algorithms [66].

## 4. Experimental Design

In this section, the experimental framework of this research study is presented. We also highlight some of the relevant issues that were considered while conducting this study.

## 4.1 Baseline Classifier Selection

It has already been established in several literature that all the classifiers are more or less affected by the class imbalance [54,56]. The performance of the classifiers on imbalanced data depends largely on how the data is modified to reduce the class imbalance. The objective of this study is to analyze and compare those data processing techniques that are used to modify the data. In that regard, all the experiments are conducted with the same classifier using the same hyperparameters for all different sampling techniques. As for the classifier, the Random Forest (RF) model was selected. RF is a powerful and robust ensemble algorithm that provides a more generalizable performance. It usually outperforms other classifiers and is less susceptible to noise and outliers. The 'RandomForestClassifier' with default parameters from the scikit-learn library was utilized in this study [67].

## 4.2   Sampling Techniques Selection

There exists a wide variety of approaches to address the class imbalance issue. Due to the significance of this problem, researchers have developed numerous methods to deal with imbalanced data. Analyzing all those approaches is impractical for a single study. On that account, we focused on the most popular approaches in imbalanced learning. Data sampling and cost-sensitive learning are by far the most common practices when dealing with skewed data. The sampling approaches can be implemented in two ways — as a data preprocessing step or by incorporating them into the ensemble learning framework. Both approaches have been proven to be quite successful in many imbalanced learning tasks [58,68–70]. The cost-sensitive learning approach can be implemented with other classifiers as well. However, it has already been established in several literature that there is not a significant difference in performance between sampling and cost-sensitive learning approaches [63,45]. Moreover, the design of cost-sensitive models is quite different than sampling approaches and requires a separate study on its own. In this regard, cost-sensitive learning approaches are not considered in this study.

This study emphasizes the analysis of sampling techniques in the imbalanced domain. Sampling as a data-preprocessing step or in the ensemble-learning framework, both types are considered. Again, there are numerous sampling techniques proposed by researchers over the years. The most popular and unique ones are utilized in this study for analysis. A total of 26 different techniques were analyzed in this study. Among them, there are 5 oversampling strategies, 10 undersampling strategies, 4 hybrid strategies, and 7 ensemble-based strategies. The ensemble strategies include bagging, boosting, and hybrid approaches. All the techniques utilized in this study are discussed in Section 5 and depicted in Figure 1.

There are several other techniques that are very similar to the ones utilized in this study. Those techniques only slightly differ in performance, and they are analogous to the ones utilized in this study in terms of methodology. Therefore, we utilized the most representative ones in the experiment. Insight into other approaches can be obtained by analyzing the similar ones used in this study.

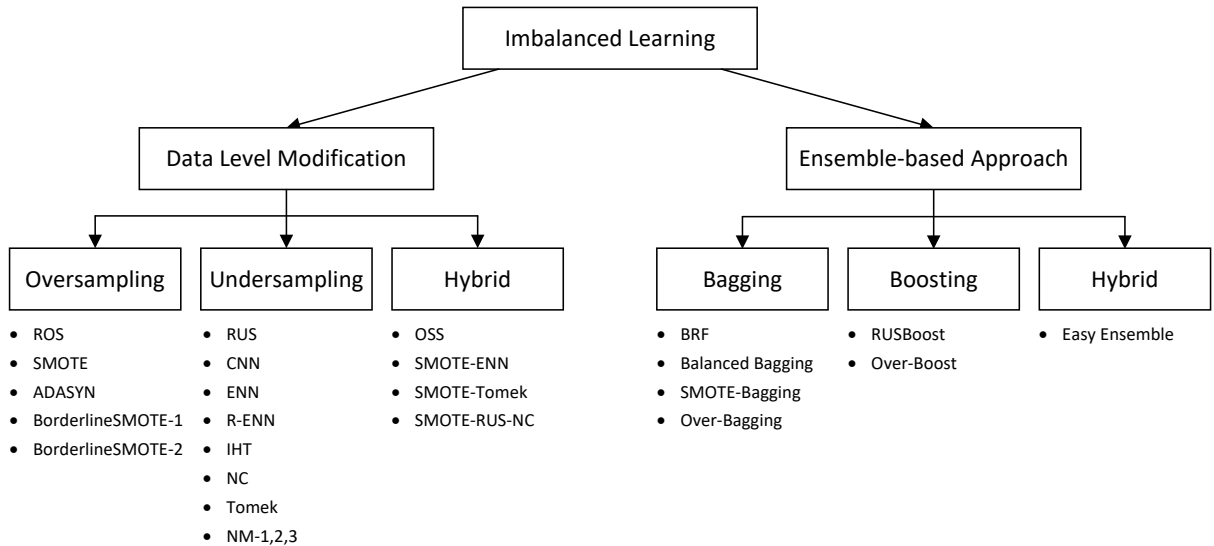

Figure 1: Methods utilized in this study

## 4.3 Dataset Selection

To thoroughly assess the performance of different techniques, they need to be tested on a wide range of datasets. Different datasets have different intrinsic characteristics, and they can influence the outcome of the sampling approaches [56]. To ensure diversity, the datasets utilized in this study were collected from different sources with a varied range of imbalance ratios. Rigorous experiments were conducted on around 60 imbalanced datasets. However, in several datasets, especially in datasets with a smaller imbalance ratio, the standard RF classifier achieved a g-mean score of greater than 99% without any data preprocessing. In such datasets, applying any preprocessing method is unnecessary. An effective comparison among various sampling techniques cannot be achieved from those datasets. Therefore, results from those datasets are not included in this manuscript. The results from the 50 remaining datasets are reported here. The imbalance ratio of these datasets varied from 1.82 to 129.44. All the datasets are binary classification datasets and all of them are publicly available in the UCI machine learning repository [71] and KEEL dataset repository [72].

A summary of the datasets is provided in the additional files [Additional File 1].

## 4.4 Experimental Framework

To avoid data leakage as well as data shift, a k-fold stratified cross-validation strategy was utilized to split the datasets into training and testing folds. The value of 'k' was chosen as 5. This value was selected to ensure that the testing set contains a fair number of minority class samples. With a value of 10, the number of samples in the minority class (especially in highly imbalanced data) might be too low in some datasets to obtain acceptable results. The sampling techniques were employed on the training sets and the results were collected from the testing sets. The dataset was normalized prior to the application of sampling techniques. The results from five different test folds were averaged to obtain the result from one iteration. The entire process was repeated 10 times with the 'random_state' parameter set to random values on each run. The results from all the iterations were then averaged to obtain the final results. The standard deviation was also calculated to study the variance of the methods. Five different measures – accuracy, sensitivity, specificity, g-mean, and roc-auc score – were obtained to assess the performance.

The outline of the proposed framework is illustrated in Figure 2.

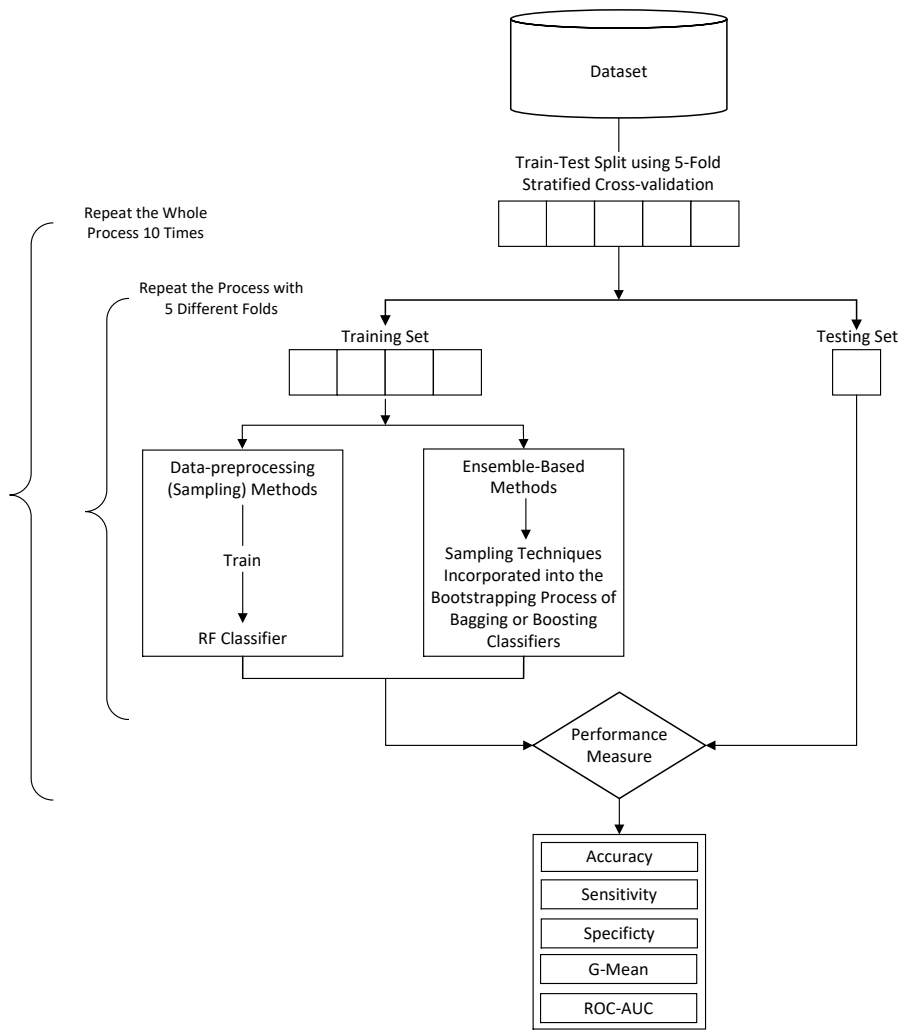

Figure 2: Outline of the experimental framework

## 5. Methods

A total of 26 different sampling techniques were analyzed in this study. They can be divided into four categories. These strategies are summarized in this section.

### 5.1 Oversampling

Oversampling refers to generating new minority class samples to balance the dataset. The oversampling techniques utilized in this study are as follows.

- **Random Oversampling (ROS)**: ROS is a simple non-heuristic approach to resample the data. It randomly selects some minority class samples from the original dataset and generates duplicates of them to balance the dataset. How many samples to be duplicated in this way can be controlled by the 'sampling ratio' ($\alpha_{os}$) parameter. Generally, samples are generated in such a way that the modified dataset contains an equal distribution between the positive and negative instances.

$$sampling\ ratio, \alpha_{os} = \frac{no.\ of\ samples\ in\ the\ minority\ class\ after\ resampling}{no.\ of\ samples\ in\ the\ majority\ class} \quad (1)$$

- **SMOTE**: SMOTE is one of the most renowned heuristic oversampling techniques used in imbalanced domain. In contrast with ROS, instead of just duplicating already existing samples, SMOTE generates synthetic samples to balance the dataset. First, it randomly selects minority class samples in the data. Then their k-nearest neighbors are identified. Finally, a synthetic minority class sample is generated along the line segments connecting the original sample with its nearest neighbors. The value of 'k' was chosen as 5 in this study.
- **ADASYN**: To overcome the limitations of the SMOTE algorithm, many different variations of SMOTE have been proposed by researchers [57]. Adaptive Synthetic Sampling or ADASYN, is one of the most popular among those variations. Instead of generating samples randomly from all minority class samples, it generates synthetic samples based on the data distribution. More samples are generated from the minority class samples that are more difficult to learn. This way, ADASYN algorithm adaptively modifies the decision boundary using data intrinsic characteristics.
- **Borderline-SMOTE**: Another popular extension of SMOTE is Borderline-SMOTE. The notion behind this algorithm is that boundary samples are more difficult to learn and more likely to be misclassified. Samples that are far away from the border are far less likely to be misclassified and hence do not need that much attention. Therefore, unlike SMOTE which generates synthetic samples using all minority class instances, Borderline-SMOTE creates new samples using only the instances that are located near the decision boundary. There are two variations of this algorithm, termed Borderline-SMOTE-1 (BS-1) and Borderline-SMOTE-2 (BS-2).

## 5.2 Undersampling

Undersampling refers to eliminating some samples from the majority class to balance the dataset. The undersampling techniques utilized in this study are as follows.

- **Random Undersampling (RUS)**: Similar to ROS, RUS is a simple non-heuristic approach to resample the data. It randomly selects the majority class samples in the data for removal. The number of samples to be eliminated can be controlled by the 'sampling ratio' ($\alpha_{us}$) parameter.

$$sampling\ ratio, \alpha_{us} = \frac{no.\ of\ samples\ in\ the\ minority\ class}{no.\ of\ samples\ in\ the\ majority\ class\ after\ resampling} \quad (2)$$

- **Condensed Nearest Neighbor (CNN):** CNN is one of the classical heuristics undersampling methods. It targets the majority class instances that lie far from the decision boundary. CNN algorithm considers them less relevant for the learning process and therefore eliminates them from the majority class. It utilizes the nearest neighbor approach to identify the distant examples. The number of samples eliminated by the CNN algorithm depends on the data. So, CNN algorithm usually reduces the imbalance ratio but cannot balance the class distribution completely.
- **Edited Nearest Neighbor (ENN)**: ENN is another classical method of undersampling. It uses the nearest neighbor rule to eliminate majority class instances from the data. If any sample is misclassified by the majority of its k-nearest neighbors, then that sample is removed from the data. This way, ENN removes noisy samples from the data. The 'k' value for ENN was taken as 3 in this study. The number of samples eliminated by the ENN algorithm depends on the data.
- **Repeated Edited Nearest Neighbor (R-ENN)**: This is similar to the ENN approach. The only difference is that here the elimination process is repeated several times to obtain a more in-depth cleaning. The number of samples eliminated in this process is larger than the ENN algorithm. This helps in reducing the imbalance ratio further.
- **Instance Hardness Threshold (IHT)**: Understanding which instances are misclassified and why they are misclassified can provide insight into data intrinsic characteristics. Smith et al. [19] developed a way to calculate this difficulty of classifying instances and termed it "Instance Hardness". They presented several ways of measuring hardness in their article. Using this hardness measure of different majority class samples, some instances can be filtered to reduce the class imbalance.

- **Tomek-link (TL)**: Tomek links are pairs of instances of opposite classes which are their own nearest neighbors. In undersampling, whenever a tomek-link is found in the data, the instance belonging to the majority class is removed from the dataset. This way, class imbalance is reduced.
- **Near Miss (NM)**: This algorithm targets the majority class samples that lie close to the minority class instances. It has several variations termed NM-1, NM-2, and NM-3. NM-1 selects the majority class samples whose mean distances to the three nearest minority class samples are the smallest. NM-2 selects the majority class samples whose mean distances to the three most distant minority class samples are the smallest. NM-3 selects the majority class samples that are closest to the minority class samples. These selected majority class samples are eliminated from the data to obtain balance.
- **Neighborhood Cleaning Rule (NC)**: NC is a modification of the original ENN approach. Here, for each sample in the dataset, its k-nearest neighbors are identified. Now, if the original sample belongs to the minority class and is misclassified by its nearest neighbors, then the majority class instances in those nearest neighbors are removed from the data. On the other hand, if the original sample belongs to the majority class and is misclassified by its nearest neighbors, then that sample is removed from the dataset. A value of 'k' was taken as 3 in the experiment.

## 5.3 Hybrid Sampling

The hybrid sampling technique refers to a combination of different sampling techniques. It can be a combination between an oversampling and an undersampling technique or a combination of two or more oversampling or undersampling techniques. The hybrid sampling techniques utilized in this study are as follows.

- **SMOTE-ENN**: This is a combination of two popular sampling techniques – SMOTE and ENN. First, ENN algorithm eliminates some majority class samples from the data. Then, SMOTE generates synthetic samples to balance the dataset.
- **SMOTE-Tomek**: This is similar to SMOTE-ENN. The only difference is that a tomek-link-based undersampling strategy is used here.
- **One-Sided Selection (OSS)**: This is a combination of two undersampling approaches. Some majority class samples are first removed based on Tomek-links. Then the majority class samples that are far from the decision boundary are removed using the ENN algorithm.

- **SMOTE-RUS-NC (SR-NC)**: This is a combination of two undersampling approaches with an oversampling technique. First, NC algorithm is utilized to reduce the class imbalance. It cannot balance the dataset. Then, RUS is utilized to randomly remove some majority class samples from the data to lower the imbalance further down to around 50%. Finally, SMOTE is utilized to generate new samples to balance the dataset.

## 5.4 Ensemble-Based Approach

This refers to the techniques that incorporate sampling strategies into the ensemble learning framework. Ensemble algorithms usually outperform other traditional machine learning algorithms like SVM, KNN, or DT. Bagging and boosting are two of the most popular ensemble approaches. While bagging attempts to reduce the variance of the learning algorithms, boosting adjusts the weights of the classifiers incrementally and adaptively to improve the model prediction. However, these ensemble algorithms remain susceptible to the class imbalance problem. To successfully realize the model's potential, the data on which they are trained needs to be modified to address the class imbalance issue. Different sampling techniques like SMOTE or RUS can easily be integrated into the learning framework of the ensemble approaches and reduce the bias due to the imbalance.

The bagging techniques utilize the concept of bootstrap aggregating [73]. Bootstrapping refers to randomly drawing samples with replacement from the original data to constitute new bootstrap replicas of the data. Several base classifiers are then trained on these different bootstrap samples, not the original data. Since the classifiers are no longer trained on the same data, rather on a subset of the original data, it introduces diversity in the training process. Finally, the predictions from different base classifiers are compiled using a majority voting or a weighted-majority voting approach. This way, the bagging ensemble classifier reduces the variance of the weak learners (base classifiers) and provides more robust performance. The boosting algorithm is aimed at building a strong classifier using a consecutive set of weak learners [74]. The first weak learner is trained on the entire dataset, while the second weak learner attempts to correct the errors of the first learner. The process is continued in this way by cascading more and more weak learners. At each stage, the weights are modified by increasing the weights of the wrongly classified instances. This enables the algorithm to learn difficult examples as well, making it a powerful technique. There are many variations of the boosting algorithm. AdaBoost, Gradient Boosting Machine (GBM), LightGBM, and XGBoost are some of the most widely used boosting techniques.

The ensemble-based approaches for imbalanced learning utilized in this study are as follows:

- **Balanced Random Forest (BRF)**: In this approach, the RF classifier construction is utilized with DTs as base learners. RF uses the same concept as the bagging technique. The difference is that in RF, each tree is trained on a random subset of the features. This makes the algorithm more robust. The number of trees used to create the forest is 100. In the bootstrapping process, the RUS algorithm is used to balance each bootstrap sample. These balanced bootstrap samples will then be used for training the base classifiers.
- **RUSBoost Classifier**: This algorithm uses the AdaBoost classifier as the boosting framework. At each iteration of the boosting framework, the data is balanced using the RUS algorithm. Decision trees are used as the base estimator. A total of 50 estimators are used in this study to form the RUSBoost classifier.
- **Balanced-Bagging Classifier (BB)**: This is directly based on the bagging technique. During the bootstrapping process, the bootstrap samples are balanced using the RUS algorithm. The base classifiers are then trained on the balanced subsets of the data. DTs are used as the base estimators, and the number of estimators used to form the classifier is 10. There are many variations of this technique with a slight difference. For instance, Exactly Balanced Bagging, Roughly Balanced Bagging, Random Balance Bagging, etc.
- **SMOTE-Bagging**: This is the same as the BB algorithm above, with the only difference of using SMOTE to balance the dataset instead of RUS.
- **Over-Bagging**: Similar to BB with the only difference of using the ROS technique to balance the dataset instead of RUS.
- **Over-Boost**: Similar to RUSBoost with the only difference of using the ROS technique to balance the dataset instead of RUS.
- **Easy Ensemble Classifier**: This is a hybrid ensemble algorithm. It is a bagging ensemble of AdaBoost classifiers trained on different balanced bootstrap samples. The balance is achieved using the RUS algorithm. The number of AdaBoost classifiers utilized in this study to form the ensemble is 10.

## 6. Results

A rigorous experiment with 26 different sampling strategies on 50 imbalanced datasets was conducted for analogy. Five different measures – accuracy, sensitivity, specificity, g-mean, and roc-auc were obtained. As it has been already discussed in Section 3, that measures like accuracy, sensitivity, and specificity cannot represent the actual performance entirely. Therefore, compound metrics like g-mean and roc-auc scores are utilized for analysis and comparison. The g-mean scores obtained from different approaches are reported in Table 2 to 4. All the tables are sorted in terms of the imbalance ratio of the datasets.

The other four measures are provided in additional files [Additional File 2].

Table 2: G-mean scores obtained (in percentage) using the standard RF classifier and after applying Over-sampling and Hybrid sampling strategies

| Dataset Name | Imbalance Ratio | RF | Oversampling ||||| Hybrid |||
| --- | --- | --- | --- | --- | --- | --- | --- | --- | --- | --- |
| | | | ROS | SMOTE | ADASYN | BorderlineSMOTE-1 | BorderlineSMOTE-2 | OSS | SMOTE-ENN | SMOTE-Tomek | SMOTE-RUS-NC |
| glass1 | 1.82 | 75.18 | 76.53 | 76.02 | 76.58 | 76.40 | 76.00 | 74.71 | 69.49 | 74.87 | 67.07 |
| ecoli-0_vs_1 | 1.86 | 98.38 | 98.23 | 97.95 | 96.93 | 97.64 | 97.57 | 98.65 | 98.65 | 98.16 | 97.98 |
| wisconsin | 1.86 | 96.05 | 96.34 | 96.26 | 96.44 | 96.25 | 96.56 | 96.27 | 96.63 | 96.44 | 97.52 |
| pima | 1.87 | 71.03 | 72.82 | 73.49 | 73.39 | 73.19 | 73.15 | 72.52 | 75.64 | 73.32 | 78.67 |
| glass0 | 2.06 | 85.92 | 85.92 | 86.06 | 84.70 | 83.24 | 83.28 | 85.03 | 80.97 | 85.59 | 79.75 |
| hf | 2.11 | 63.37 | 65.14 | 65.54 | 66.15 | 64.84 | 64.99 | 67.65 | 63.59 | 66.41 | 66.19 |
| yeast1 | 2.46 | 63.54 | 68.16 | 69.06 | 69.05 | 68.67 | 69.14 | 67.07 | 70.87 | 69.29 | 70.46 |
| vehicle2 | 2.88 | 98.26 | 98.52 | 98.63 | 98.74 | 98.70 | 98.70 | 98.44 | 98.35 | 98.56 | 98.55 |
| vehicle1 | 2.9 | 64.38 | 72.01 | 73.90 | 75.55 | 75.07 | 76.09 | 71.19 | 78.64 | 74.41 | 79.10 |
| vehicle3 | 2.99 | 59.65 | 67.48 | 69.89 | 71.14 | 70.80 | 72.11 | 66.22 | 77.82 | 70.34 | 75.32 |
| vehicle0 | 3.25 | 96.22 | 96.84 | 96.80 | 96.48 | 96.07 | 96.42 | 96.30 | 95.56 | 96.49 | 95.45 |
| new-thyroid1 | 5.14 | 96.06 | 93.27 | 93.80 | 94.71 | 94.62 | 96.40 | 95.40 | 94.03 | 93.73 | 97.45 |
| ecoli2 | 5.46 | 82.77 | 88.56 | 88.45 | 87.67 | 86.55 | 86.09 | 85.28 | 88.96 | 88.32 | 88.38 |
| glass6 | 6.38 | 88.39 | 91.65 | 93.31 | 93.24 | 93.42 | 92.05 | 92.22 | 91.78 | 92.78 | 90.52 |
| yeast3 | 8.1 | 82.17 | 88.27 | 88.92 | 90.14 | 89.73 | 90.44 | 84.41 | 91.86 | 88.83 | 93.30 |
| ecoli3 | 8.6 | 70.56 | 72.96 | 81.22 | 83.08 | 75.37 | 82.24 | 72.14 | 86.37 | 81.89 | 86.91 |
| page-blocks0 | 8.79 | 87.14 | 88.87 | 90.70 | 91.08 | 90.65 | 90.30 | 87.90 | 91.66 | 91.07 | 93.19 |
| yeast-2_vs_4 | 9.08 | 85.52 | 85.25 | 87.98 | 87.68 | 86.80 | 88.76 | 85.39 | 91.50 | 88.78 | 91.33 |

| Dataset | | | | | | | | | | | |
|---|---|---|---|---|---|---|---|---|---|---|---|
| yeast-0-2-5-6_vs_3-7-8-9 | 9.14 | 62.03 | 65.97 | 68.98 | 71.43 | 67.94 | 65.95 | 62.67 | 71.09 | 69.04 | 71.37 |
| Vowel0 | 9.98 | 74.95 | 76.57 | 80.83 | 75.42 | 80.91 | 82.92 | 75.14 | 81.28 | 81.02 | 88.16 |
| led7digit-0-2-4-5-6-7-8-9_vs_1 | 10.97 | 83.44 | 84.49 | 84.83 | 85.29 | 85.32 | 84.77 | 83.96 | 60.73 | 84.94 | 82.49 |
| glass2 | 11.59 | 7.00 | 31.30 | 35.56 | 34.23 | 32.34 | 37.98 | 13.88 | 49.06 | 34.82 | 60.51 |
| ecoli-0-1-4-7_vs_5-6 | 12.28 | 64.87 | 80.37 | 82.62 | 83.04 | 81.73 | 82.93 | 71.13 | 81.68 | 82.70 | 81.43 |
| glass4 | 15.47 | 73.44 | 80.26 | 83.25 | 80.49 | 84.55 | 90.52 | 73.31 | 81.99 | 83.83 | 81.47 |
| ecoli4 | 15.8 | 78.77 | 87.67 | 88.26 | 83.39 | 81.28 | 89.74 | 80.95 | 87.00 | 88.76 | 87.71 |
| page-blocks-1-3_vs_4 | 15.86 | 84.59 | 91.98 | 97.04 | 97.23 | 97.69 | 99.18 | 94.33 | 95.35 | 97.22 | 95.71 |
| abalone9-18 | 16.4 | 33.72 | 38.44 | 65.62 | 62.60 | 59.58 | 66.05 | 34.50 | 66.69 | 64.64 | 72.95 |
| yeast-1-4-5-8_vs_7 | 22.1 | 0.00 | 11.39 | 34.08 | 36.27 | 25.67 | 32.56 | 0.00 | 48.61 | 33.19 | 60.44 |
| yeast | 23.15 | 82.43 | 88.56 | 89.32 | 90.53 | 90.14 | 90.27 | 84.21 | 91.69 | 89.03 | 93.04 |
| flaref | 23.79 | 28.86 | 29.52 | 48.66 | 47.51 | 41.94 | 40.45 | 52.88 | 77.37 | 51.98 | 82.26 |
| yeast4 | 28.1 | 30.74 | 36.70 | 64.84 | 64.81 | 58.87 | 66.38 | 32.72 | 73.34 | 65.22 | 82.80 |
| winequality-red-4 | 29.17 | 3.87 | 2.41 | 37.36 | 38.07 | 22.39 | 32.12 | 3.26 | 52.12 | 36.11 | 64.44 |
| yeast-1-2-8-9_vs_7 | 30.57 | 25.65 | 26.38 | 39.81 | 39.08 | 31.42 | 31.00 | 19.46 | 51.05 | 39.64 | 68.18 |
| yeast5 | 32.73 | 70.68 | 83.23 | 91.17 | 91.39 | 91.54 | 89.78 | 72.74 | 94.84 | 90.69 | 96.42 |
| winequality-red-8_vs_6 | 35.44 | 0.00 | 0.00 | 19.08 | 16.00 | 14.48 | 10.26 | 0.00 | 49.57 | 20.08 | 69.52 |
| ecoli_013vs26 | 39.14 | 31.15 | 33.93 | 0.00 | 0.00 | 23.33 | 10.00 | 35.17 | 33.33 | 0.00 | 48.21 |
| abalone-17_vs_7-8-9-10 | 39.31 | 21.32 | 24.48 | 59.87 | 59.54 | 53.54 | 61.01 | 20.35 | 72.14 | 58.28 | 81.55 |
| yeast6 | 41.4 | 49.20 | 72.29 | 76.22 | 75.57 | 70.42 | 73.53 | 54.11 | 79.29 | 76.43 | 86.06 |
| winequality-white-3_vs_7 | 44 | 19.97 | 10.00 | 35.58 | 33.00 | 20.26 | 26.98 | 29.94 | 36.75 | 31.95 | 66.01 |
| winequality-red-8_vs_6-7 | 46.5 | 20.00 | 20.00 | 33.72 | 33.67 | 24.05 | 20.69 | 20.00 | 34.95 | 33.71 | 60.12 |
| kddcup-land_vs_portsweep | 49.52 | 97.87 | 97.89 | 97.89 | 100.00 | 97.89 | 97.89 | 97.53 | 97.89 | 97.89 | 98.59 |
| abalone-19_vs_10-11-12-13 | 49.69 | 0.00 | 6.52 | 15.68 | 17.86 | 8.13 | 12.96 | 0.00 | 43.77 | 15.45 | 62.11 |
| winequality-white-3-9_vs_5 | 58.28 | 8.94 | 0.89 | 19.59 | 16.92 | 15.15 | 19.24 | 8.94 | 33.58 | 20.47 | 68.04 |
| poker-8-9_vs_6 | 58.4 | 8.05 | 10.19 | 47.04 | 65.10 | 51.37 | 54.30 | 8.94 | 47.52 | 45.33 | 79.51 |
| winequality-red-3_vs_5 | 68.1 | 0.00 | 0.00 | 0.00 | 0.00 | 0.00 | 4.24 | 0.00 | 23.84 | 0.00 | 47.27 |

| Dataset Name | | | | | | | | | | |
|---|---|---|---|---|---|---|---|---|---|---|
| abalone-20_vs_8-9-10 | 72.69 | 8.27 | 34.78 | 56.00 | 56.00 | 41.05 | 70.40 | 8.33 | 69.60 | 55.99 | 81.54 |
| kddcup-land_vs_satan | 75.67 | 97.32 | 97.32 | 97.32 | 100.00 | 97.32 | 97.32 | 98.80 | 97.32 | 97.32 | 97.11 |
| poker-8-9_vs_5 | 82 | 0.00 | 0.00 | 5.35 | 6.25 | 0.00 | 0.00 | 0.00 | 7.11 | 4.46 | 47.54 |
| poker-8_vs_6 | 85.88 | 0.00 | 11.55 | 62.96 | 67.83 | 48.47 | 58.99 | 0.00 | 63.44 | 62.01 | 74.92 |
| abalone19 | 129.44 | 0.00 | 0.00 | 22.44 | 18.63 | 8.93 | 15.39 | 0.00 | 38.92 | 18.61 | 70.14 |

Table 3: G-mean scores obtained (in percentage) after applying Undersampling strategies

| Dataset Name | Imbalance Ratio | RUS | CNN | ENN | R-ENN | IHT | NC | Tomek | NM-1 | NM-2 | NM-3 |
|---|---|---|---|---|---|---|---|---|---|---|---|
| glass1 | 1.82 | 75.08 | 74.05 | 69.66 | 67.87 | 68.55 | 72.42 | 74.55 | 59.94 | 65.14 | 73.79 |
| ecoli-0_vs_1 | 1.86 | 98.05 | 96.39 | 98.27 | 97.95 | 96.62 | 98.37 | 98.38 | 98.34 | 98.41 | 98.55 |
| wisconsin | 1.86 | 97.17 | 96.87 | 97.33 | 97.38 | 96.69 | 97.17 | 96.16 | 96.62 | 96.17 | 96.17 |
| pima | 1.87 | 74.54 | 73.61 | 74.30 | 72.25 | 72.57 | 74.24 | 72.35 | 72.29 | 67.68 | 72.29 |
| glass0 | 2.06 | 85.25 | 86.78 | 81.22 | 76.60 | 76.32 | 80.68 | 83.73 | 84.17 | 84.98 | 84.32 |
| hf | 2.11 | 69.57 | 66.86 | 70.82 | 66.64 | 69.74 | 64.97 | 67.18 | 68.09 | 64.29 | 71.63 |
| yeast1 | 2.46 | 70.72 | 70.28 | 70.91 | 64.25 | 65.64 | 71.42 | 67.57 | 57.98 | 55.47 | 68.60 |
| vehicle2 | 2.88 | 98.00 | 97.99 | 98.35 | 97.71 | 90.64 | 98.36 | 98.41 | 98.46 | 75.01 | 98.46 |
| vehicle1 | 2.9 | 77.99 | 76.88 | 79.31 | 75.62 | 70.72 | 79.35 | 70.95 | 74.46 | 63.94 | 74.27 |
| vehicle3 | 2.99 | 75.90 | 71.46 | 77.32 | 76.67 | 71.21 | 76.80 | 66.22 | 68.39 | 62.79 | 68.47 |
| vehicle0 | 3.25 | 95.71 | 97.19 | 95.37 | 94.27 | 88.18 | 95.57 | 96.08 | 88.35 | 93.39 | 93.74 |
| new-thyroid1 | 5.14 | 98.10 | 96.00 | 95.86 | 95.81 | 96.15 | 96.17 | 96.06 | 96.57 | 95.40 | 97.48 |
| ecoli2 | 5.46 | 88.46 | 88.27 | 88.78 | 88.85 | 86.75 | 88.45 | 85.97 | 81.92 | 83.81 | 85.46 |
| glass6 | 6.38 | 89.38 | 89.59 | 91.84 | 92.02 | 88.41 | 92.66 | 91.69 | 89.84 | 92.15 | 91.17 |
| yeast3 | 8.1 | 92.48 | 90.58 | 88.77 | 89.80 | 91.17 | 89.58 | 84.84 | 82.95 | 79.44 | 87.09 |
| ecoli3 | 8.6 | 86.50 | 76.25 | 83.26 | 84.01 | 84.31 | 80.76 | 72.46 | 41.18 | 66.70 | 77.96 |
| page-blocks0 | 8.79 | 93.57 | 91.55 | 88.69 | 89.95 | 93.92 | 89.38 | 87.60 | 67.83 | 40.91 | 90.15 |

| Dataset | IR | | | | | | | | | | |
|---|---|---|---|---|---|---|---|---|---|---|---|
| yeast-2_vs_4 | 9.08 | 91.82 | 87.98 | 89.10 | 89.43 | 92.03 | 90.21 | 86.43 | 84.48 | 81.44 | 89.17 |
| yeast-0-2-5-6_vs_3-7-8-9 | 9.14 | 70.12 | 69.51 | 72.52 | 71.85 | 66.81 | 69.50 | 62.70 | 53.27 | 45.85 | 65.19 |
| Vowel0 | 9.98 | 90.18 | 93.65 | 75.52 | 75.52 | 89.14 | 74.74 | 74.95 | 86.60 | 80.86 | 88.35 |
| led7digit-0-2-4-5-6-7-8-9_vs_1 | 10.97 | 81.13 | 85.43 | 83.47 | 82.61 | 82.85 | 83.50 | 83.44 | 43.70 | 22.09 | 83.05 |
| glass2 | 11.59 | 67.93 | 31.39 | 15.79 | 19.55 | 66.26 | 17.78 | 9.92 | 49.11 | 44.44 | 50.15 |
| ecoli-0-1-4-7_vs_5-6 | 12.28 | 83.02 | 84.35 | 70.61 | 70.29 | 82.68 | 75.35 | 67.38 | 64.00 | 81.24 | 82.76 |
| glass4 | 15.47 | 80.90 | 79.49 | 76.39 | 79.09 | 82.20 | 80.19 | 76.96 | 84.52 | 53.81 | 85.46 |
| ecoli4 | 15.8 | 86.72 | 87.36 | 77.54 | 77.15 | 87.70 | 77.51 | 75.87 | 84.53 | 80.10 | 86.25 |
| page-blocks-1-3_vs_4 | 15.86 | 92.58 | 98.27 | 83.67 | 83.67 | 93.13 | 85.89 | 84.59 | 87.39 | 87.33 | 97.09 |
| abalone9-18 | 16.4 | 71.80 | 55.06 | 38.39 | 41.53 | 72.17 | 43.90 | 35.99 | 58.09 | 46.28 | 65.17 |
| yeast-1-4-5-8_vs_7 | 22.1 | 60.53 | 11.97 | 5.72 | 10.11 | 60.07 | 1.63 | 0.82 | 47.33 | 33.30 | 45.79 |
| yeast | 23.15 | 92.28 | 88.98 | 88.96 | 89.87 | 90.70 | 89.33 | 84.53 | 82.94 | 80.28 | 87.00 |
| flaref | 23.79 | 81.71 | 50.35 | 71.75 | 74.25 | 78.76 | 52.40 | 50.57 | 31.87 | 20.73 | 67.45 |
| yeast4 | 28.1 | 83.05 | 46.21 | 51.71 | 56.81 | 83.72 | 49.11 | 33.34 | 56.29 | 52.63 | 70.27 |
| winequality-red-4 | 29.17 | 63.77 | 15.58 | 13.41 | 21.86 | 60.33 | 3.67 | 3.86 | 35.31 | 46.99 | 63.41 |
| yeast-1-2-8-9_vs_7 | 30.57 | 66.13 | 37.22 | 30.98 | 32.20 | 63.98 | 22.48 | 22.89 | 56.03 | 39.68 | 58.08 |
| yeast5 | 32.73 | 95.72 | 86.39 | 84.79 | 86.72 | 95.21 | 82.91 | 73.18 | 85.27 | 83.11 | 88.73 |
| winequality-red-8_vs_6 | 35.44 | 73.11 | 14.21 | 11.13 | 9.98 | 58.74 | 9.00 | 0.00 | 52.36 | 61.11 | 62.61 |
| ecoli_013vs26 | 39.14 | 62.57 | 51.82 | 50.06 | 51.19 | 56.64 | 52.59 | 31.71 | 61.90 | 44.79 | 50.60 |
| abalone-17_vs_7-8-9-10 | 39.31 | 80.58 | 39.50 | 21.91 | 22.62 | 67.12 | 27.35 | 20.38 | 56.04 | 24.70 | 39.68 |
| yeast6 | 41.4 | 85.25 | 65.83 | 65.54 | 66.95 | 84.24 | 58.34 | 53.48 | 45.53 | 68.29 | 77.49 |
| winequality-white-3_vs_7 | 44 | 69.48 | 73.08 | 20.97 | 20.97 | 59.78 | 21.39 | 19.97 | 64.07 | 37.94 | 55.00 |
| winequality-red-8_vs_6-7 | 46.5 | 55.59 | 23.28 | 20.00 | 20.00 | 39.00 | 20.00 | 20.00 | 44.68 | 42.10 | 55.24 |
| kddcup-land_vs_portsweep | 49.52 | 98.57 | 97.42 | 97.87 | 97.87 | 97.17 | 97.87 | 97.87 | 97.82 | 97.79 | 97.79 |
| abalone-19_vs_10-11-12-13 | 49.69 | 60.47 | 0.00 | 0.00 | 0.00 | 13.03 | 0.00 | 0.00 | 43.34 | 41.59 | 57.76 |
| winequality-white-3-9_vs_5 | 58.28 | 65.30 | 19.10 | 8.93 | 8.93 | 46.26 | 8.94 | 8.94 | 44.67 | 45.05 | 52.21 |

| Dataset Name | | | | | | | | | | | |
|---|---|---|---|---|---|---|---|---|---|---|---|
| poker-8-9_vs_6 | 58.4 | 72.76 | 49.45 | 8.94 | 8.05 | 17.21 | 8.05 | 6.26 | 50.26 | 43.24 | 67.79 |
| winequality-red-3_vs_5 | 68.1 | 51.19 | 0.00 | 0.00 | 0.00 | 34.02 | 1.41 | 0.00 | 13.61 | 53.91 | 50.32 |
| abalone-20_vs_8-9-10 | 72.69 | 79.71 | 33.77 | 18.69 | 19.99 | 49.95 | 18.16 | 11.69 | 64.36 | 32.81 | 38.41 |
| kddcup-land_vs_satan | 75.67 | 96.96 | 99.94 | 97.32 | 97.32 | 97.32 | 97.32 | 97.32 | 99.22 | 95.86 | 99.88 |
| poker-8-9_vs_5 | 82 | 54.40 | 2.98 | 0.00 | 0.00 | 0.00 | 0.00 | 0.00 | 45.63 | 47.55 | 49.27 |
| poker-8_vs_6 | 85.88 | 63.90 | 11.55 | 0.00 | 0.00 | 5.77 | 0.00 | 0.00 | 43.83 | 42.96 | 59.06 |
| abalone19 | 129.44 | 70.63 | 0.00 | 0.00 | 0.00 | 3.26 | 0.00 | 0.00 | 34.69 | 32.29 | 55.29 |

Table 4: G-mean scores obtained (in percentage) using ensemble approaches

| Dataset Name | Imbalance Ratio | BRF | Balanced Bagging | SMOTE-Bagging | Over-Bagging | RUSBoost | Over-Boost | Easy Ensemble |
|---|---|---|---|---|---|---|---|---|
| glass1 | 1.82 | 75.48 | 71.86 | 72.85 | 69.61 | 73.11 | 71.92 | 74.74 |
| ecoli-0_vs_1 | 1.86 | 98.23 | 97.50 | 97.73 | 97.80 | 98.05 | 97.13 | 97.05 |
| wisconsin | 1.86 | 97.32 | 95.44 | 94.92 | 94.53 | 93.58 | 94.40 | 96.21 |
| pima | 1.87 | 75.40 | 71.04 | 69.98 | 67.19 | 70.71 | 73.44 | 75.39 |
| glass0 | 2.06 | 85.46 | 83.55 | 82.76 | 81.93 | 77.71 | 78.82 | 81.36 |
| hf | 2.11 | 70.14 | 66.60 | 64.47 | 60.84 | 61.08 | 63.34 | 68.02 |
| yeast1 | 2.46 | 71.04 | 68.13 | 64.03 | 58.71 | 67.94 | 70.17 | 69.65 |
| vehicle2 | 2.88 | 98.21 | 96.51 | 96.29 | 95.85 | 95.48 | 97.22 | 97.34 |
| vehicle1 | 2.9 | 79.68 | 74.91 | 67.69 | 62.86 | 69.49 | 73.50 | 77.14 |
| vehicle3 | 2.99 | 77.36 | 71.69 | 64.70 | 58.68 | 70.18 | 74.13 | 74.67 |
| vehicle0 | 3.25 | 96.00 | 95.41 | 93.44 | 91.70 | 94.62 | 96.04 | 96.39 |
| new-thyroid1 | 5.14 | 98.67 | 95.04 | 93.33 | 91.90 | 97.20 | 96.75 | 96.98 |
| ecoli2 | 5.46 | 88.44 | 87.10 | 85.32 | 84.79 | 82.49 | 88.11 | 87.71 |
| glass6 | 6.38 | 91.33 | 91.16 | 91.89 | 90.19 | 90.32 | 92.30 | 89.63 |
| yeast3 | 8.1 | 93.17 | 92.47 | 86.05 | 83.48 | 84.80 | 90.68 | 91.26 |
| ecoli3 | 8.6 | 87.20 | 86.17 | 72.37 | 64.42 | 73.41 | 71.33 | 86.31 |
| page-blocks0 | 8.79 | 93.72 | 91.74 | 88.42 | 85.63 | 83.45 | 90.00 | 92.14 |
| yeast-2_vs_4 | 9.08 | 92.28 | 93.30 | 86.70 | 82.20 | 82.53 | 83.74 | 92.82 |
| yeast-0-2-5-6_vs_3-7-8-9 | 9.14 | 71.35 | 73.91 | 63.07 | 60.64 | 61.35 | 71.13 | 70.72 |
| Vowel0 | 9.98 | 89.32 | 92.56 | 86.10 | 86.84 | 80.35 | 83.05 | 92.39 |
| led7digit-0-2-4-5-6-7-8-9_vs_1 | 10.97 | 80.25 | 81.32 | 82.24 | 81.99 | 80.99 | 81.67 | 74.60 |
| glass2 | 11.59 | 75.17 | 57.94 | 27.58 | 21.91 | 39.33 | 43.06 | 72.39 |
| ecoli-0-1-4-7_vs_5-6 | 12.28 | 83.05 | 78.51 | 76.38 | 74.70 | 78.27 | 80.17 | 79.83 |
| glass4 | 15.47 | 81.42 | 79.14 | 70.14 | 72.69 | 78.83 | 78.35 | 79.67 |
| ecoli4 | 15.8 | 87.52 | 81.67 | 82.36 | 79.99 | 87.76 | 81.77 | 85.56 |
| page-blocks-1-3_vs_4 | 15.86 | 94.03 | 94.98 | 89.89 | 84.83 | 95.11 | 96.37 | 94.46 |
| abalone9-18 | 16.4 | 73.82 | 69.07 | 54.55 | 37.67 | 57.17 | 60.67 | 74.34 |
| yeast-1-4-5-8_vs_7 | 22.1 | 63.88 | 57.88 | 6.46 | 11.40 | 48.22 | 56.23 | 60.25 |
| yeast | 23.15 | 93.15 | 92.27 | 85.80 | 83.19 | 86.11 | 90.72 | 90.83 |
| flaref | 23.79 | 83.06 | 74.83 | 36.74 | 31.31 | 64.13 | 58.61 | 73.96 |
| yeast4 | 28.1 | 84.97 | 81.79 | 56.19 | 31.22 | 68.79 | 70.72 | 81.97 |
| winequality-red-4 | 29.17 | 65.42 | 63.63 | 31.40 | 7.41 | 46.24 | 52.55 | 64.28 |
| yeast-1-2-8-9_vs_7 | 30.57 | 69.04 | 64.80 | 22.35 | 15.18 | 43.61 | 50.87 | 66.45 |
| yeast5 | 32.73 | 95.98 | 95.80 | 84.27 | 77.34 | 90.05 | 86.41 | 95.22 |
| winequality-red-8_vs_6 | 35.44 | 78.80 | 66.87 | 9.26 | 0.99 | 40.94 | 28.70 | 78.03 |
| ecoli_013vs26 | 39.14 | 61.72 | 46.50 | 0 | 10.99 | 38.80 | 29.30 | 56.00 |
| abalone-17_vs_7-8-9-10 | 39.31 | 82.77 | 80.37 | 43.20 | 20.57 | 66.09 | 73.00 | 82.68 |
| yeast6 | 41.4 | 86.40 | 83.95 | 71.91 | 68.43 | 78.58 | 71.81 | 83.46 |
| winequality-white-3_vs_7 | 44 | 70.57 | 63.75 | 11.63 | 2.00 | 42.79 | 58.83 | 74.72 |
| winequality-red-8_vs_6-7 | 46.5 | 60.73 | 56.20 | 25.23 | 19.41 | 45.46 | 36.64 | 64.00 |
| kddcup-land_vs_portsweep | 49.52 | 98.81 | 98.08 | 97.38 | 97.38 | 98.41 | 97.04 | 99.05 |
| abalone-19_vs_10-11-12-13 | 49.69 | 65.09 | 59.69 | 12.35 | 0.82 | 49.81 | 46.31 | 67.33 |
| winequality-white-3-9_vs_5 | 58.28 | 68.84 | 59.69 | 5.37 | 0.00 | 37.73 | 30.13 | 67.45 |
| poker-8-9_vs_6 | 58.4 | 78.07 | 60.09 | 33.28 | 0.00 | 32.82 | 33.33 | 41.07 |
| winequality-red-3_vs_5 | 68.1 | 49.37 | 46.23 | 0.00 | 0.00 | 16.26 | 2.81 | 47.76 |
| abalone-20_vs_8-9-10 | 72.69 | 83.62 | 79.68 | 46.10 | 18.73 | 72.03 | 60.45 | 80.97 |
| kddcup-land_vs_satan | 75.67 | 98.51 | 98.93 | 96.90 | 96.90 | 95.45 | 96.18 | 98.73 |
| poker-8-9_vs_5 | 82 | 55.02 | 50.31 | 3.58 | 0.00 | 33.43 | 39.76 | 35.70 |
| poker-8_vs_6 | 85.88 | 68.98 | 57.87 | 25.73 | 0.00 | 31.07 | 35.68 | 50.20 |
| abalone19 | 129.44 | 75.09 | 68.16 | 2.43 | 0.00 | 53.60 | 50.79 | 73.67 |

## 7. Discussion

In this section, we study the results obtained from the experimentation and provide a meticulous analysis of different sampling techniques. We examine their behavior on a wide range of datasets and identify their advantages and limitations. Moreover, we present a comparative analysis among these techniques and provide recommendations on their usage.

### 7.1 Effect of sampling techniques on data distribution

We start by looking into the effect of sampling techniques on the distribution of data. The goal of the sampling algorithms is to obtain a balance between the classes in the data so that the standard classification algorithms can effectively classify positive and negative cases without getting biased towards the majority class. However, the data does not necessarily have to be equally balanced for the classification algorithms to work properly. Many of the sampling techniques have a parameter called "sampling_ratio" with which the level of balance that we want to obtain can be controlled. The common approach is to equalize the two classes, and this has been followed during the experimentation. Other heuristic sampling techniques remove or generate samples based on certain criteria, which depend on the data distribution. These techniques do not balance the dataset. Depending on the type of algorithm, the dataset might remain quite imbalanced even after sampling with some of these techniques.

For simplicity of analysis, we divide these algorithms into two categories—the ones that balance the class distribution and the ones that do not. They are presented in Table 5.

Table 5: Categorization of algorithms based on their effect on data distribution

| Category-01: | Oversampling | Undersampling | Hybrid | Ensemble |
|---|---|---|---|---|
| Sampling techniques that can balance the dataset | ▪ ROS<br>▪ SMOTE<br>▪ ADASYN<br>▪ Borderline-SMOTE-1<br>▪ Borderline-SMOTE-2 | ▪ RUS<br>▪ NM-1, 2, 3 | ▪ SMOTE-Tomek<br>▪ SMOTE-RUS-NC | ▪ BRF<br>▪ RUSBoost<br>▪ Over-Boost<br>▪ Balanced-Bagging<br>▪ SMOTE-Bagging<br>▪ Over-Bagging<br>▪ Easy Ensemble |
| **Category-02:** | **Undersampling** | **Hybrid Sampling** | | |
| Sampling techniques that do not balance the dataset | ▪ CNN<br>▪ ENN<br>▪ R-ENN<br>▪ IHT<br>▪ NC<br>▪ Tomek-link | ▪ OSS<br>▪ SMOTE-ENN | | |

Category–01 techniques can balance the dataset by generating or removing the necessary number of samples. However, this approach has a major drawback. When the imbalance ratio is large, these techniques have to generate or eliminate a considerable number of samples to balance the dataset. For instance, in the poker-8-9_vs_5 dataset, there are 2049 samples in the majority class with only 25 samples in the minority class. To balance the dataset using sampling techniques, more than 2000 samples need to be created or removed. Duplicating so many samples from only 25 original samples does not improve the performance. Generating so many augmented samples from only 25 samples using SMOTE-based techniques leads to a loss of generalization. Generating these many samples obscures the decision boundary increasing the occurrences of overlapping between the classes. This ultimately leads to poor performance.

On the other hand, randomly eliminating samples from the data can cause a loss of information. In the poker-8-9_vs_5 dataset, using RUS results in a distribution of only 25 samples in both positive and negative classes. Lowering the imbalance ratio this way by removing so many samples is quite impractical. A model trained on such data tends to perform very poorly in real-world scenarios when it is tested on an external dataset that comes from a different distribution. As a significant amount of information is lost, the model loses its discriminating capability.

Bagging-based ensemble algorithms follow the same process by generating or removing any number of samples to balance the data. However, those are an ensemble of many (100 in this study) base classifiers, each trained on a balanced bootstrap sample. The bootstrap samples are a subset of the original data, which are then balanced using oversampling (ROS or SMOTE) or undersampling (RUS). When balancing each bootstrap sample, there is a loss of information or the generation of too many samples. However, the effect of these is mitigated to some extent by the use of 100 different bootstrap samples, each with a slightly different data distribution. Some information might be lost in one bootstrap sample due to undersampling, but that information may be present in several other bootstrap samples. The results from all the trees trained on different bootstrap samples are then combined to obtain the final result. Consequently, ensemble algorithms usually produce comparatively better results than the sampling techniques (discussed in more detail in section 7.5).

As for the techniques of category–02, they use different heuristics to eliminate samples from the majority class. But the resultant data is not balanced. While some of these techniques considerably reduce the imbalance ratio, others do not. As a result, the modified dataset can still remain quite imbalanced even after resampling. This is quite apparent in techniques like Tomek-link, ENN, R-ENN, NC, and OSS. These techniques remove only a small number of samples from the majority class. The distribution of positive and negative samples before and after sampling is presented in additional files (Table 7.1). For instance, in the ecoli2 (IR = 5.46) dataset, the number of samples in the majority and minority classes is 283 and 52, respectively. After resampling, the number of samples remaining in the majority class is 278 for Tomek-link, 256 for ENN, 245 for R-ENN, 260 for NC, and 271 for OSS. So, it is quite evident that these undersampling techniques are not quite efficient in removing many samples. However, these techniques remove samples very carefully so as not to lose important information. Consequently, some of these techniques like ENN, R-ENN, and NC produce comparable performance, sometimes even better than other popular techniques like SMOTE or RUSBoost.

Among the techniques in category – 02, CNN removes the maximum number of samples from the majority class. On the ecoli2 dataset mentioned above, the number of samples remaining in the majority class after CNN is applied is only 48, which is even lower than the number of positive samples. IHT, on the other hand, carefully removes many samples from the majority class by utilizing data intrinsic characteristics. It has proven quite effective in terms of performance as well.

The effect of sampling techniques (category – 02) on data distribution of a sample dataset 'winequality-red-8_vs_6' is illustrated in Figure 3. It presents the imbalance ratio of the data before and after sampling.

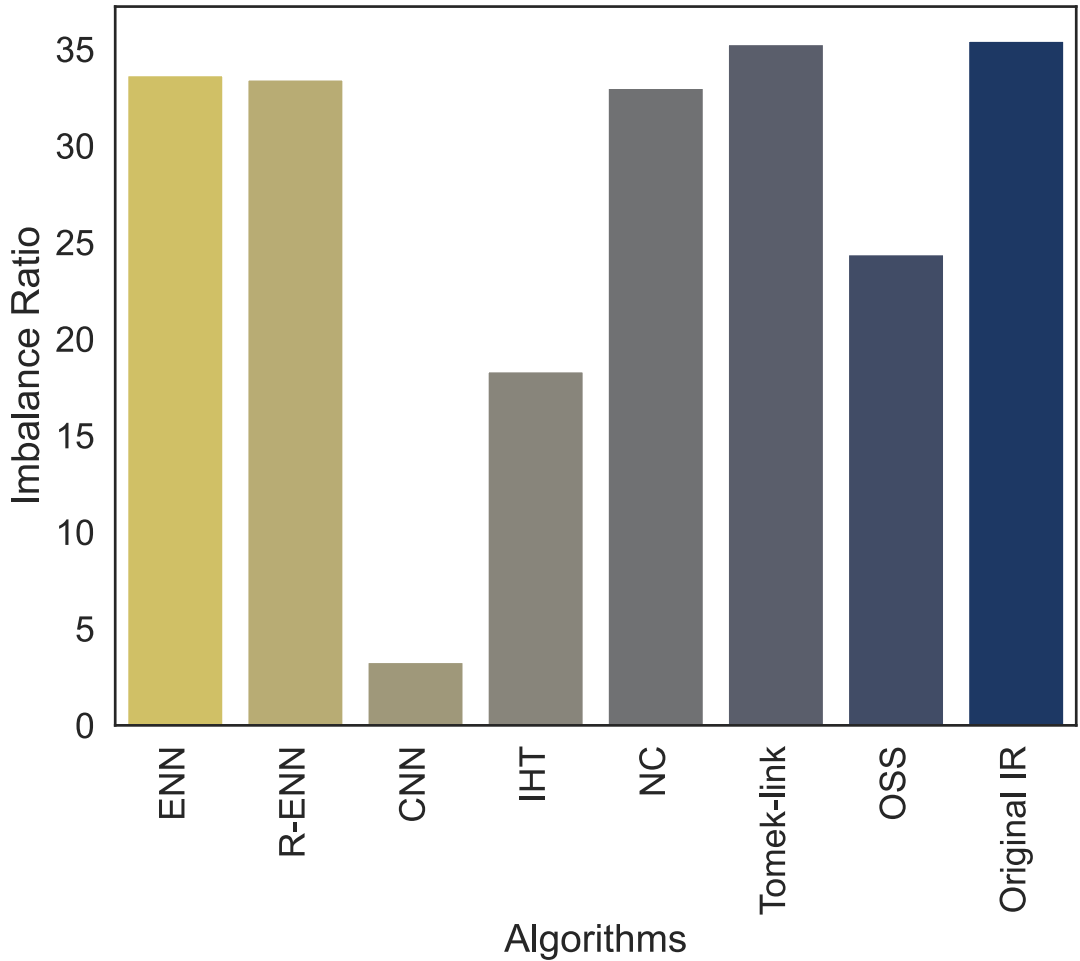

Figure 3: Original IR vs IR after sampling using category-02 approaches for 'winequality-red-8_vs_6' dataset

## 7.2  Effect of imbalance ratio on the overall performance

It has been observed during the experiment that the Imbalance Ratio (IR) of the data has a significant effect on the efficacy of the sampling techniques. How different sampling techniques modify the data distribution plays an important role in that effect. In this section, we want to provide a comprehensive discussion of the impact of IR on the efficacy of the sampling techniques. We want to establish a generic methodology for the selection of sampling techniques based on IR by analyzing the comparative analysis of their performance.

To obtain a thorough evaluation, 50 datasets with varying degrees of imbalance were utilized in this study. The IR of the datasets varied from 1.82 to 130. The performance of the standard RF classifier is considered the baseline. The g-mean score is considered for comparing the performances. The g-mean score obtained from different sampling techniques is presented in Table 2 to 4.

As can be observed from Table 2, the standard RF classifier works very well when the IR is too small. Resampling the data in scenarios like that does not provide much benefit. There is only a slight improvement in performance from the application of some resampling techniques, particularly the oversampling techniques. However, in the case of the undersampling techniques, the performance drops a little. Some valuable information might be lost when eliminating samples from the data. Therefore, using undersampling algorithms in small imbalance scenarios is futile and should be avoided. Standard classification algorithms like RF can handle small imbalance and provide comparable performance.

As we move towards the datasets with higher IR, the improvement in performance from the application of sampling techniques starts to become more apparent. For instance, in the yeast3 dataset (IR = 8.1), RF achieves a g-mean score of 82.17%. Resampling the data with SMOTE boosts the performance to 88.92%. All the sampling techniques are usually able to improve the performance when the data is somewhat imbalanced and there is no significant difference between different approaches like oversampling, undersampling, and ensemble. However, as we move towards datasets with a considerable amount of imbalance (IR>10), the difference becomes apparent.

In datasets with high imbalance, the RF classifier usually performs very poorly. In some cases, the g-mean score is 0, indicating a complete failure in the classification task. A similar scenario can be observed for any other classification algorithm. The classifier gets completely biased by the majority class and is unable to distinguish the positive cases. In scenarios like these, increasing the number of minority class instances using oversampling techniques might seem like the logical approach. However, in practice, this is not the case, as can be observed from the results. The oversampling techniques only provide a small improvement in performance compared to other approaches. This is due to the fact that when the IR is large, the oversampling techniques produce a significant number of synthetic samples from a very small number of actual samples. As a result, the model loses its generalizability and performs poorly during validation. This also holds for oversampling-based ensemble algorithms like Over-Bagging or SMOTE-

Bagging. They hardly provide any improvement over SMOTE or other oversampling techniques. As for different oversampling techniques, ROS is the worst. Merely duplicating samples does not provide much benefit. SMOTE is definitely better than ROS. SMOTE and its different variations tend to provide similar performance. In a few cases, variations like Borderline-SMOTE or ADASYN are able to outperform SMOTE, but that is not dependent on IR. They are dependent on other data intrinsic characteristics like class separability as they use data density or borderline examples to generate new samples. The difference in performance among these variations of SMOTE techniques is usually negligible and is not investigated further.

While ROS performs very poorly in datasets with higher imbalance, RUS surprisingly provides much better performance. It easily outperforms other oversampling techniques as well as most other undersampling techniques. It has already been discussed in the previous section that most of the heuristic undersampling techniques, except IHT and CNN, are unable to remove a good number of samples from the majority class and reduce the imbalance in high IR datasets. Thereby, those techniques perform poorly on such datasets. IHT and CNN, however, have shown to produce comparable results to RUS. There are some complications with using RUS. RUS removes examples from the dataset arbitrarily. This makes the algorithm quite unstable, producing different results on different iterations. This is a high-variance technique, which is discussed in detail in the following section (Section 7.3). Moreover, removing data can cause loss of valuable information. The problem intensifies with a larger class imbalance as a significant number of samples are removed. Ensemble algorithms are quite useful in alleviating the problem and providing a more generalized performance. Algorithms like BRF or Easy-ensemble, which use RUS while training base learners, usually outperform simple RUS-based preprocessing approaches. In datasets with a smaller imbalance, however, algorithms like SMOTE or other undersampling approaches provide comparable performance to RUS and can be more effective in handling class imbalance.

As for the hybrid sampling techniques, tomek-link based methods like SMOTE-Tomek or OSS are not very effective. As can be observed from 'Additional File 3', the number of samples eliminated from the majority class using Tomek-links is quite limited. Therefore, the data remains quite imbalanced in OSS-based sampling or requires the generation of a huge number of samples using SMOTE in SMOTE-Tomek, both leading to poor performance. SMOTE-ENN is definitely an improvement over SMOTE and works quite well in small to mid-imbalance ranges. Nonetheless, the performance is inferior compared to RUS or BRF in high imbalance scenarios. SMOTE-RUS-NC is a clever hybridization of three sampling techniques, and it has shown to outperform other sampling techniques, including RUS. The idea behind it is to reduce the loss of information or generation of too many samples by first reducing the imbalance using a heuristic undersampling method NC. Then obtain an optimal balance iteratively using a combination of SMOTE and RUS. Although NC is not very effective in reducing the imbalance (Additional File 3), some other heuristic approaches can be utilized in its place to bring down the IR first and then use SMOTE and RUS in conjunction. This kind of hybridization can be quite effective in handling data with different degrees of imbalance and is more promising than using the algorithms independently.

## 7.3   Stability analysis of the sampling techniques

Machine learning algorithms are generally stochastic in nature. They explicitly use randomness in the learning process, resulting in slightly different outcomes from different iterations. Randomness is also present in the sampling process. Non-heuristic approaches like ROS and RUS are entirely random, while other approaches like SMOTE also use a random selection of examples to generate synthetic samples. The problem with these stochastic processes is that the obtained predictions might vary with different iterations. This is often undesirable, and it is expected that there should not be significant variations in the obtained results. To properly assess the variations in the predicted outcomes, the experiments were conducted 10 times. The final results are the average of the 10 runs. The standard deviation in the results (g-mean score) for different approaches is also calculated and provided in the 'Additional file 4'.

The variation depends not only on the strategies but also on the data. As the data is more imbalanced, the variations in the results are much more significant. Using sampling techniques to reduce class distribution usually lowers the variance, but that is not the case for all sampling techniques. For some cases, the results fluctuate significantly. Some of the techniques are highly variant while still providing very good g-mean scores. This is unacceptable as there is no guarantee of a good outcome. Especially in real-world applications, this is quite unfavorable and perilous. An appropriate algorithm should provide better consistent performance with a small variance in results in different runs.

To provide a broad overview and comparison of the variance of different strategies, the average of the standard deviations in the results of each algorithm for all 50 datasets was calculated. For ease of analysis, we divide the algorithms into three categories: low variance, medium variance, and high variance. They are presented in Table 6. One thing to point out here is that this categorization is strictly based on the average results obtained from the 50 datasets included in this study. The purpose of the table below is to offer a broad overview of the stability of various techniques. The algorithms are arranged in ascending order according to their average variance.

Table 6: Categorization of algorithms based on variance

| Low Variance | Mid Variance | High Variance |
|---|---|---|
| - NM3 | - Borderline SMOTE-1 | - SMOTE-ENN |
| - BRF | - Easy Ensemble | - Over-Boost |
| - IHT | - ENN | - RUS |
| - ROS | - Tomek | - OSS |
| - NM1 | - SMOTE | - CNN |
| - NM2 | - NC | - SMOTE-RUS-NC |
| - ADASYN | - Borderline SMOTE-2 | - Over-Bagging |
| - R-ENN | - SMOTE-Tomek | - Balanced-Bagging |
| | | - SMOTE-Bagging |
| | | - RUSBoost |

One thing to consider here is that an algorithm might be a low variance strategy, but its performance is comparatively inferior to other algorithms, making it unsuitable. ROS is one such algorithm. It usually provides very poor performance, but it is a low-variance strategy. RUS, on the other hand, is a high-variance strategy, but it usually produces much better results than many other techniques. In spite of the outstanding performance, it is not feasible to use such a high-variance strategy on highly imbalanced datasets. The same goes for other techniques like RUSBoost, which is found to be the algorithm with the highest variance. One way to alleviate the problem is to use the bagging approach to reduce the variance. This has been found to be quite effective in not only reducing the variance but also improving the performance. The BRF classifier is a perfect example of a low-variance, high-performance strategy. It is one of the algorithms with the lowest variances while also being the best performing classifier. However, it is not true for other bagging-based approaches like Over-Bagging or SMOTE-Bagging. Besides being high-variance strategies, they perform very poorly on data with mid to high-range IR. Balanced Bagging uses a combination of oversampling and undersampling techniques. Although it is a high-variance strategy, it provides comparable performance to other algorithms.

Figure 4 presents a subtle illustration of the average performance of the algorithms and their respective variance.

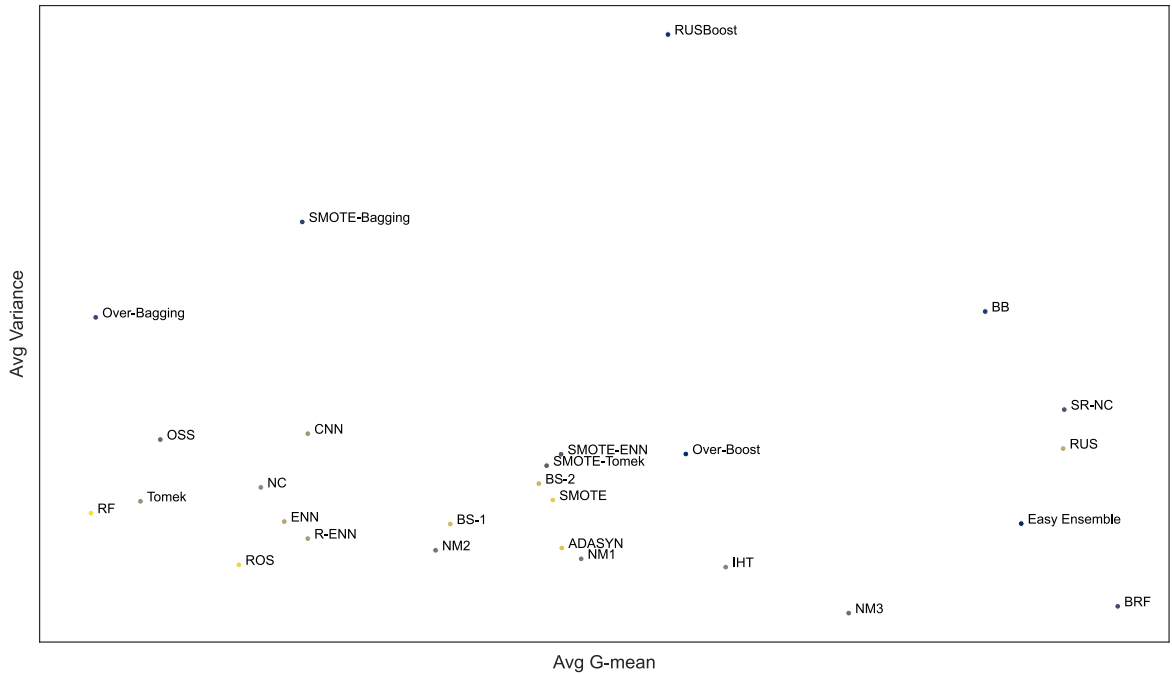

**Figure 4**: Average variance vs avg. g-mean score of different sampling strategies

In the x-axis of figure 3, we have the avg. g-mean which is the average of the g-mean scores obtained in all 50 datasets for a particular algorithm. However, other factors like imbalance ratio or sample size are not considered. Therefore, the average performance does not necessarily represent the superiority of a particular algorithm. But it is useful to establish a comparative analogy among different techniques.

BRF and Easy Ensemble are clearly two of the best performing strategies in imbalanced domain. They also have limited variance. SMOTE and its variations are mid-level variance strategies, and they provide decent performance. Algorithms like OSS, CNN, NC, ENN, R-ENN, and Tomek-link provide comparatively inferior performance while also showing mid to high-range deviation in performance.

## 7.4   Time complexity of different sampling techniques

The time it requires to train a particular algorithm depends on the dataset size, dimensionality, and preprocessing strategy. Data sampling usually requires a small amount of time compared to other preprocessing techniques like feature selection. However, in the case of very large datasets, it needs to be taken into consideration. This section provides a comparison of the time required by various sampling strategies. The time that was required to perform each preprocessing technique and train the algorithm for all the datasets is provided in the 'Additional File 5'. For comparison, we look at the average time required by different algorithms and, based on that, we group the algorithms into three categories presented in Table 7.

Table 7: Categorization of algorithms based on their time complexity

| Low | Mid | High |
|---|---|---|
| - RUS | - BRF | - Easy ensemble |
| - RUSBoost | - SMOTE-Bagging | - IHT |
| - ROS | - SMOTE | - CNN |
| - Over-Boost | - NC | |
| - BB | - ADASYN | |
| - Over-Bagging | - Borderline SMOTE-1 | |
| - NM1, NM2, NM3 | - SMOTE-ENN | |
| - Tomek-link | - SMOTE-Tomek | |
| - OSS | - SMOTE-RUS-NC | |
| - ENN | - Borderline SMOTE-2 | |
| - R-ENN | | |

CNN requires the maximum time to eliminate samples from the dataset, followed by IHT. Both are heuristic undersampling approaches that are quite successful in eliminating samples from the data and provide decent performance. Non-heuristic approaches like ROS and RUS require the smallest amount of time. The ensemble algorithms that are based on those techniques also require little time. The time required by the ensemble algorithms depends significantly on the number of base estimators used to form the ensemble. For instance, BRF uses 100 base estimators, making it more time-consuming compared to Balanced Bagging, which uses only 10. Easy ensemble takes a comparatively longer time to train as it is an ensemble of ensembles (AdaBoost learner).

## 7.5 Performance comparison of different ensemble-based approaches in imbalanced learning

In this research, we studied the performance of six ensemble approaches belonging to three different categories. Although there are several other algorithms proposed in the literature, they are analogous to the ones employed in the experiment. Therefore, insight into their performance can be obtained by analyzing the approaches presented in this study.

The ensemble approaches used in the imbalanced domain are basically based on two different techniques—bagging and boosting. In the bagging-based techniques, each bootstrap subset of the data is resampled, and the DTs are trained on different samples. The results from all different DTs are aggregated, forming the ensemble. Consequently, the bagging-based approaches are more robust in performance. In boosting techniques, the data is balanced by some sampling technique in each iteration of the boosting algorithm. The boosting framework attempts to reduce the bias. While there are many different boosting algorithms, AdaBoost is the most widely used approach and the one utilized in this study. Although these two approaches are quite different in terms of methodology, a significant difference in their performance in imbalanced learning is caused by the resampling approach utilized to balance the dataset.

The ensemble approaches (Over-Bagging, SMOTE-Bagging) where oversampling techniques like ROS or SMOTE are used to balance the data, tend to provide the poorest performance. On average, directly processing the data using the SMOTE algorithm and training an RF classifier on the resampled data usually produces a better result than approaches like SMOTE-Bagging. The reason behind this can be speculated as follows: while bootstrapping, the number of minority class instances available on each bootstrap sample is further reduced. Therefore, the information available on minority class to produce new samples is limited. Generating new samples from a smaller number of original samples affects the quality of the data on which the classifier is trained. This harms the classifier's ability to distinguish minority class samples from the majority class, reducing the model's performance.

The problem is alleviated when undersampling techniques (RUS) are used. This approach is followed in the BRF, or Balanced-Bagging classifier. In these cases, the bootstrap samples are balanced by removing a necessary number of majority class samples. This can cause a loss of information. However, as a large number of base learners trained on a different bootstrap subset of the data are used to form the bagging ensemble, the loss of information is quite minimized. Consequently, BRF and BB are two of the highest performing classifiers over a range of datasets.

The complications with the RUS approach become more apparent when it is used in the boosting framework. The RUSBoost classifier is found to be the strategy with the highest variance. The decisions from this classifier vary significantly over different iterations. This is due to the fact that, in the boosting framework, the entire dataset is directly balanced by randomly eliminating samples, leading to a significant loss of information. Bagging, on the other hand, uses bootstrap sampling to reduce variance, making it much more suitable for imbalanced classification.

Easy Ensemble is a hybridization of bagging and boosting. It uses a bag of boosted learners for classification. Each AdaBoost classifier is trained on a different balanced bootstrap sample. Balancing is achieved using the RUS algorithm. The predictions from different boosted learners are then aggregated, forming the easy ensemble. This hybrid approach attempts to exploit the advantages of both bagging and boosting algorithms, making it one of the highest performing classifiers. The only drawback is that this approach is quite time-consuming compared to others.

In summary, bagging-based approaches are more effective in handling imbalances compared to boosting. Especially in the case of the highly imbalanced datasets, their performance is unparallel. Undersampling is much more effective than oversampling when incorporated into the ensemble framework. The BRF classifier is found to be the most effective and highest performing approach with the lowest variance, followed by the easy ensemble classifier.

## 7.6 Performance analysis of different resampling techniques in imbalanced learning

Data preprocessing using resampling techniques is the most popular approach in imbalanced learning. Sampling techniques can be divided into three categories: oversampling, undersampling, and hybrid sampling. There are many different approaches found in the literature, and here in this study, we have experimented with 19 popular algorithms. After preprocessing the data with these techniques, an RF classifier was used to obtain the results. However, any other classification algorithm can be utilized with the resampled dataset. In this section, we have provided a comparative analysis of these techniques, and discussed their advantages and limitations.

*7.6.1 Oversampling:* The non-heuristic oversampling approach ROS is found to be one of the poorest performing approaches. It merely duplicates the already existing minority class samples, which does not add any new information and only increases the complexity. Generating identical samples this way to balance the dataset is not enough to shift the bias, resulting in poor performance. Although this is rarely used directly, it is found to be incorporated with the ensemble framework in different algorithms like Over-Bagging, Over-Boosting, or Under-Over-Bagging. However, these algorithms also tend to provide poor performance. This is apparent from the sensitivity scores provided in 'Additional File 2'. In many datasets, the sensitivity scores obtained from these algorithms are very poor (<10%) indicating the fact that the ROS approach is not capable of shifting the bias from the majority class.

SMOTE is quite an improvement to ROS. It is a more sophisticated approach, generating new synthetic samples using interpolation. There are many different extensions of this algorithm, three of which are tested in this study. SMOTE works pretty well as long as there exists a good number of minority class samples in the data. When the number of minority class instances is small, the algorithm starts to overfit the data, lowering the performance. Depending on other data characteristics like density or overlapping subclasses, the other variations of the SMOTE algorithm like ADASYN or Borderline-SMOTE produce better performance. However, those algorithms also suffer from similar limitations like SMOTE. SMOTE is also incorporated in ensemble frameworks like SMOTE-Bagging or SMOTE-Boosting. However, these algorithms tend to perform poorly compared to the SMOTE approach (reasons are discussed in section 7.5). As a thumb rule, we can hypothesize from the experimental results that the SMOTE algorithm works well when there are a good number of minority class examples and the class imbalance is lower. In low IR scenarios, this is one of the best-performing techniques.

*7.6.2 Undersampling:* Non-heuristic undersampling approach RUS has shown better performance than many other approaches. However, this has a major downside. This is a high-variant strategy. The inconsistency in results is quite undesirable. The problem intensifies in real-world scenarios when it is tested on external data, making it less suitable for practical applications. The problem is alleviated when it is incorporated into the ensemble-learning framework. BRF and Easy Ensemble are two of the highest performing classifiers, both of which use RUS to balance the bootstrap samples. Therefore, it can be inferred that the RUS algorithm should be merged with the bagging framework to obtain desirable, consistent performance.

The effectiveness of the heuristic undersampling techniques relies significantly on their ability to eliminate noisy and unimportant majority class samples from the data and reduce the class imbalance. Some of the classical heuristic approaches like ENN, R-ENN, NC, or Tomek-link are not very good at removing examples. They can eliminate only a limited number of samples from the data (discussed in section 7.1), making them ineffective in mid-to high IR scenarios. However, in a small imbalance scenario, they can be quite effective as they specifically target the majority class examples that misclassify the minority class instances. This way, they remove some of the borderline or noisy examples from the data, improving the overall performance. Since these algorithms are unable to bring down the IR on their own, they can be more effective when merged with other oversampling techniques or used in combination with other undersampling approaches (discussed later in this section).

Near Miss is another classical heuristic undersampling technique. This has three different variations (NM-1, NM-2, and NM-3), each of which utilizes a slightly different approach to reducing imbalance. The NM-3 approach is found to be the most effective in handling imbalance and is one of the best performing undersampling techniques with the smallest variance. IHT is another sophisticated undersampling technique that takes into consideration data complexity and other internal characteristics while eliminating samples. These two are found to be the most effective undersampling techniques, with consistently better performance than others. In many cases, they also outperformed SMOTE-based oversampling.

***7.6.3 Hybrid sampling:*** Hybrid sampling techniques use a combination of oversampling and undersampling techniques to balance the dataset. Undersampling techniques like ENN or Tomek-link, which are unable to bring down the class imbalance single-handedly, can be merged with an oversampling technique like SMOTE to obtain balance. These kinds of hybridization (SMOTE-ENN or SMOTE-Tomek) are able to produce better results than using the algorithms singly. SMOTE-ENN in particular, which uses ENN as an undersampling technique (more effective than Tomek-link), has shown to produce comparatively better results than SMOTE and its variations. SMOTE-RUS-NC, which is a hybridization of two undersampling approaches (RUS and NC) with SMOTE, has displayed even better performance than SMOTE-ENN, or RUS itself in general.

Hybrid sampling techniques present a more promising solution to imbalanced learning. Sampling techniques like SMOTE or Near-miss are good at handling imbalances when IR is comparatively lower. However, with increasing IR, these techniques are unable to produce desirable results single-handedly. Hybridization among different approaches can induce a proper balance, producing superior performance. The hybrid techniques found in the literature so far are usually just a combination of two or more oversampling and undersampling techniques. A more sophisticated approach needs to be formed, which focuses on how to intelligently hybridize sampling techniques to obtain better results. This is an open research issue which requires further investigation.

In highly imbalanced scenarios, oversampling or undersampling techniques are usually not enough to shift the bias, leading to poor performance. These techniques, when used independently, have resulted in a zero g-mean score in many datasets. This indicates the ineffectiveness of such algorithms when the IR is high. Hybrid sampling techniques like SMOTE-RUS-NC or Random Balance are more useful in such scenarios. A better approach is to integrate the sampling techniques into the ensemble learning framework. The bagging framework, in particular, aggregates the predictions from a large number of base learners, leading to more robust and less biased performance. This kind of architecture is quite useful in datasets with large imbalances, and they (BRF or Easy Ensemble classifier) have shown unparallel performance in highly imbalanced datasets. Integrating hybrid sampling techniques with the ensemble framework should be able to provide even better performance. This requires further investigation and research. One major drawback of these ensemble approaches is that they can become computationally quite expensive when other sophisticated data preprocessing tasks like wrapper methods of feature selection (Genetic algorithm, Ant-colony optimization) need to be incorporated with them. The process becomes even more complicated when the dataset is large with many different attributes. Directly preprocessing the entire data using sampling techniques and then using a particular classification algorithm along with the feature selection strategy requires a significantly less amount of time.

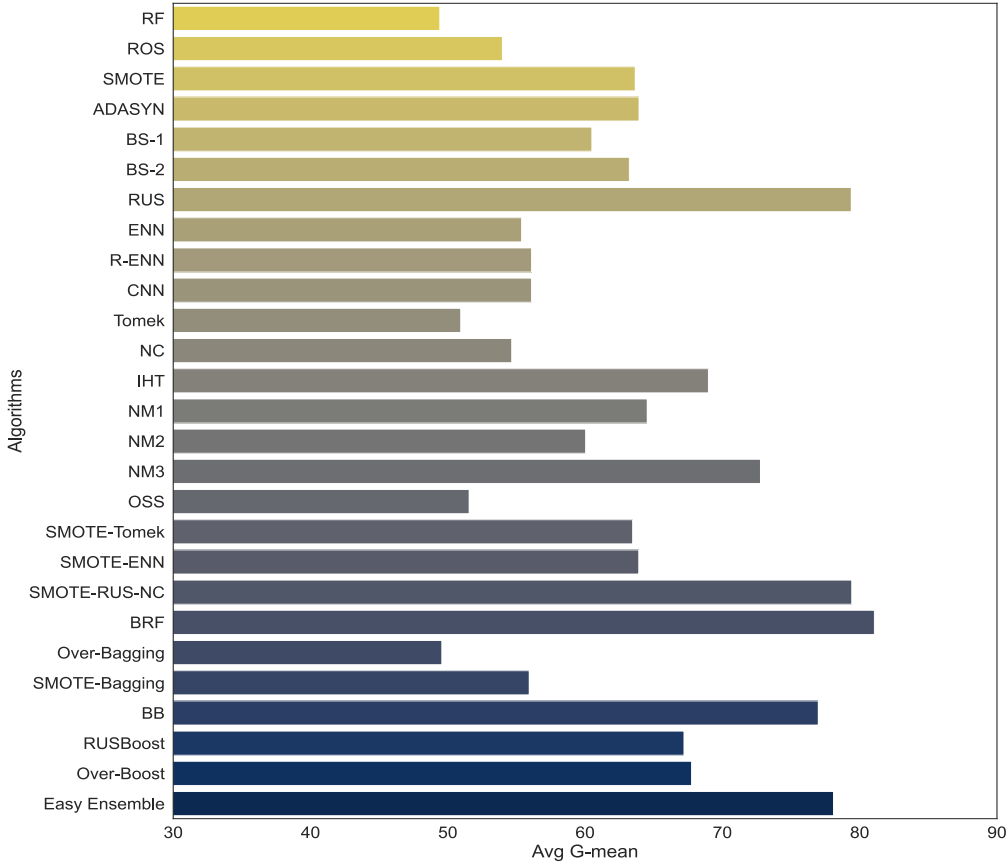

**Figure 5**: Performance comparison in terms of average g-mean score obtained from all 50 datasets

Figure 5 shows a comparison of all the techniques used in this study based on the average g-mean score from all 50 datasets to give a general idea of how well they work.

## 8. Conclusion

Learning from imbalanced data is quite challenging, but class imbalance is very prevalent in many practical applications. This has brought a lot of attention to the research community. A wide range of techniques have been proposed and they use different strategies to deal with the skewed distribution. However, not every strategy is suitable for every imbalance scenario. While they have shown greater improvement in performance in many cases, they also suffer from certain limitations. While there are a sheer number of approaches used in imbalanced learning, the questions appear: which algorithm is more suitable for a particular application; which characteristics of a particular strategy make it superior to the others; what are the drawbacks of applying a particular technique; and how to overcome those limitations. Understanding these factors can pave the way to developing more suitable and powerful algorithms to learn from imbalanced data.

This study focuses on answering such questions. A thorough investigation has been conducted into 26 different sampling algorithms in that regard. The most representative sampling techniques are chosen for experimentation. A wide variety of datasets (a total of 50) are utilized for proper validation. The advantages and limitations of different techniques, as well as a comparative analysis among the techniques, have been discussed in detail. We also looked at how the limitations of the sampling methods could be circumvented and how to choose the best sampling method for a given application. Some future research directions that can be quite beneficial to imbalanced learning have also been outlined.

There are a few issues that are not covered in detail in this research study. For instance, data dimensionality reduction techniques like Feature Selection (FS) are not considered. FS algorithms usually improve the performance. It has been established in some previous literature that feature selection algorithms work well when the data is adequately balanced [70]. How different sampling techniques are affected by different feature selection algorithms is an open research issue and requires further investigation. Moreover, since there exist a sheer number of sampling algorithms, covering all of them in a single study is difficult. Therefore, the most representative sampling techniques are utilized in this study such that some reasonable intuition about how other similar approaches would perform can be obtained. Cost-sensitive learning is another popular approach in the imbalanced domain. Since this approach is quite different than the sampling algorithms, it is not covered in this study. In future work, we plan to present a comprehensive analysis of cost-sensitive techniques and their differences with data sampling techniques.

There are a few limitations to the current study. One of the issues is the lack of availability of large imbalanced datasets. The datasets utilized here are collected from public repositories. Some of these datasets are quite small. However, to avoid bias, the minimum number of samples required to be considered in this study was taken as 200. An open access repository containing large imbalanced datasets would be quite beneficial for these kinds of research purposes. Such a repository would be able to provide a standard benchmark for evaluating various strategies in a more justified manner. Researchers from all over the globe can contribute to developing such a repository for imbalanced data analysis. Another issue is that we did not investigate in detail different data intrinsic characteristics like class-overlapping and how they affect different sampling techniques.

The focus of this study is to analyze the overall performance to provide a comparative analysis and obtain a comprehensive understanding of different algorithms used in imbalanced learning. We believe our work will provide great insight into different sampling techniques, their effectiveness, and limitations. The analogy will also pave the way to developing new, more robust sampling techniques that can overcome the limitations of the current strategies and provide better performance in different imbalance scenarios.

## Supplementary material

Supplementary data associated with this article can be found at https://github.com/newaz-aa/Empirical_analysis_of_sampling_techniques.

## Declaration of competing interest

The authors declare that they have no known competing financial interests or personal relationships that could have appeared to influence the work reported in this paper.

## Sources of funding

This research did not receive any specific grants from funding agencies in the public, commercial, or not-for-profit sectors.